\documentclass[10pt,twocolumn,letterpaper]{article}

\usepackage{iccv_2020}
\usepackage{times}
\usepackage{epsfig}
\usepackage{graphicx}
\usepackage{amsmath}
\usepackage{amssymb}
\usepackage{multirow}
\usepackage{multicol}

\usepackage{color} 
\usepackage[dvipsnames]{xcolor}
\usepackage{booktabs}
\usepackage{pifont}
\usepackage{tabularx}
\usepackage{nicematrix}
\usepackage{gensymb}
\usepackage{subcaption}

\usepackage{xspace}
\usepackage{xcolor}

\usepackage{acronym}
\acrodef{NVS}{Neural View Synthesis}
\acrodef{NeRF}{Neural Radiance Field}
\acrodef{HVS}{Human Visual System}
\acrodef{NSS}{Natural Scene Statistics}

\newcommand{\figref}[1]{Figure~\ref{fig:#1}}
\newcommand{\secref}[1]{Section~\ref{sec:#1}}

\newcommand{\tableref}[1]{Table~\ref{tab:#1}}

\newcommand{\cdms}{\,cd/m$^2$\xspace}

\newcommand{\scene}[1]{{\fontfamily{pcr}\selectfont #1}}

\newcommand{\dslab}{{\fontfamily{bch}\selectfont Lab}}
\newcommand{\dsfieldwork}{{\fontfamily{bch}\selectfont Fieldwork}}
\newcommand{\dsllff}{{\fontfamily{bch}\selectfont LLFF}}

\newcommand{\cmark}{\ding{51}}%
\newcommand{\xmark}{\ding{55}}%

\newcommand{\datasetName}{Forward-Facing Video}

\DeclareGraphicsExtensions{.pdf,.eps}

\usepackage{hyperref}
\hypersetup{breaklinks=true,bookmarks=false}
\usepackage[capitalize]{cleveref}
\iccvfinalcopy 

\def\iccvPaperID{3341} 

\ificcvfinal\fi

\begin{document}

\title{Perceptual Quality Assessment of NeRF and Neural View Synthesis Methods \\ for Front-Facing Views}

\author{
Hanxue Liang\textsuperscript{\rm 1 },
Tianhao Wu\textsuperscript{\rm 1 },
 Param Hanji\textsuperscript{\rm 1},
Francesco Banterle\textsuperscript{\rm 2},
Hongyun Gao\textsuperscript{\rm 1}, \\ 
Rafal Mantiuk\textsuperscript{\rm 1},
Cengiz Oztireli\textsuperscript{\rm 1}\\
\textsuperscript{\rm 1}University of Cambridge
\textsuperscript{\rm 2}ISTI-CNR
}

%
\maketitle

\ificcvfinal\thispagestyle{empty}\fi

\begin{abstract}
Neural view synthesis (NVS) is one of the most successful techniques for synthesizing free viewpoint videos, capable of achieving high fidelity from only a sparse set of captured images. This success has led to many variants of the techniques, each evaluated on a set of test views typically using image quality metrics such as PSNR, SSIM, or LPIPS. There has been a lack of research on how NVS methods perform with respect to perceived video quality. We present the first study on perceptual evaluation of NVS and NeRF variants. For this study, we collected two datasets of scenes captured in a controlled lab environment as well as in-the-wild. In contrast to existing datasets, these scenes come with reference video sequences, allowing us to test for temporal artifacts and subtle distortions that are easily overlooked when viewing only static images.
We measured the quality of videos synthesized by several NVS methods in a well-controlled perceptual quality assessment experiment as well as with many existing state-of-the-art image/video quality metrics. We present a detailed analysis of the results and recommendations for dataset and metric selection for NVS evaluation.


\end{abstract}

\section{Introduction}

Synthesizing photorealistic novel views of a complex scene from a sparse set of RGB images is a fundamental challenge in image-based rendering. Various representations and methods have been developed to accurately model the image formation process and handle complex geometry, materials, and lighting conditions \cite{chen1993view,shade1998layered,shum2000review,idr,diff_sdf}. More recently, \ac{NVS} via implicit representations has shown promising results. In particular, methods such as \ac{NeRF} \cite{original_nerf} and its successors \cite{mipnerf,mipnerf360,plenoxel,dirctvoxgo,ibrnet,NeX} have attracted great interest due to their outstanding fidelity and robustness. However, assessing the performance of contemporary \ac{NVS} methods is not straightforward as it is closely tied to the final applications. Specifically, \ac{NVS} methods are increasingly developed in immersive and realistic AR/VR applications, it is thus crucial for the methods to synthesize high-quality free-viewpoint videos with unnoticeable artifacts to human users.

The current protocol for comparing \ac{NVS} methods involves computing image quality metrics, such as PSNR, SSIM \cite{ssim} and LPIPS \cite{lpips}, on a subset of hold-out views for a few scenes. Even dedicated benchmarks \cite{scannerf,nerf_robust_benchmark} follow the same evaluation protocol. Since the main  objective of NVS methods is to offer interactive exploration of novel views, we argue that those methods should be evaluated on \emph{video sequences} rather than individual sparse views, ideally in a subjective quality evaluation experiment. Thus, we identify two key limitations in existing evaluation protocols. \ding{202} They rely exclusively on image quality metrics, which can be problematic because these metrics may not correlate well with subjective judgments, especially when used for a task they are not designed for~\cite{Cadik2012a,hanji2022comparison,Ponomarenko2015}. Since most of the image quality metrics have not been calibrated or validated on the distortions specific to novel view synthesis, their predictions could be too noisy to quantify perceived quality. \ding{203} The evaluation protocol lacks assessment on video sequences, which can reveal temporal artifacts and subtle distortions, such as flickering or floating ghost images, that are easily noticeable in video but difficult to spot in static images~\cite{judder,Denes2020flicker,vmaf1,fvvdp}.
This issue is compounded by the limited nature of commonly used NVS datasets, which do not have reference videos for evaluating NVS methods.

\begin{figure*} 
    \centering
    \includegraphics[width=\textwidth]{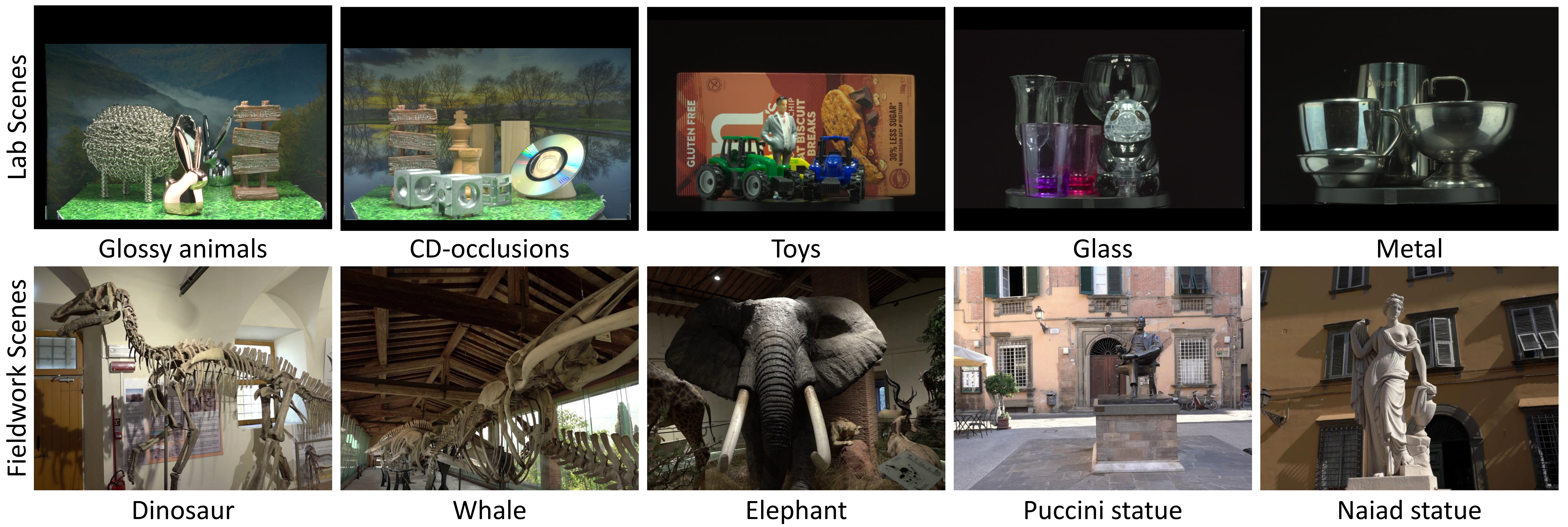}
    \caption{A subset of scenes from our collected \dslab{} (first row) and \dsfieldwork{} (second row) datasets. 
    Our datasets include both controlled laboratory and in-the-wild scenes, each with reference video sequences.
    We selected a diverse range of objects composed of various materials such as wood, marble, metal, and glass etc. 
    }\label{data-vis}
\end{figure*}

To address these problems, we first collect two new datasets: a \dslab{} dataset captured using 2D gantry in well-controlled laboratory conditions, and a \dsfieldwork{} dataset, captured in-the-wild with the help of either a gimbal or a camera slider (\secref{dataset}). Each captured scene contains several sparse training views and a reference test video intended to evaluate \ac{NVS} methods.
We use the two new datasets together with the popular \dsllff{} dataset \cite{llff} to reconstruct the video sequences by 8 NVS methods and 2 variants of generalizable NVS methods. The output videos of these methods are then evaluated by human participants in a subjective quality assessment experiment (\secref{iq-experiment}). The results of that experiment serve as ground-truth scores for testing how well the existing image and video quality metrics can predict the perceptual performance of \ac{NVS} methods (\secref{metrics}). 

In summary, the main contributions of this work are:
\begin{itemize}
    \item Two new datasets with front-facing views and video references for full-reference evaluation of synthesized videos, 
    \item A subjective quality assessment of videos synthesized by 8 NVS methods (and two generalizable NeRF variants) measured via a perceptual quality assessment experiment,
    \item An objective evaluation of existing image/video quality metrics on synthesized videos to assess how well these metrics correlate with subjective quality.
   \item A thorough analysis of metrics that elucidates the limitation in the current NVS evaluation protocol and reveals the crucial need for video assessment and video metrics. In light of this, we provide practical recommendations to enhance the efficacy of evaluation processes.
\end{itemize}
The datasets and the results of the quality assessment experiment will be made publicly available.
\section{Related Work}
\subsection{Quality assessment of NVS methods}

Most works on NVS methods and NVS benchmarks \cite{original_nerf, scannerf, llff, NeX, deepview, NSVF} evaluate on sparse hold-out views using image quality metrics. An exception is the Light Field Benchmark~\cite{light_field_bench}, where light field interpolation methods were evaluated on video sequences. On the contrary, our focus is on assessing the perceptual quality of NVS methods and evaluating how well current objective metrics can predict subjective quality. 
Such subjective benchmarks have previously motivated and advanced other areas such as tone mapping~\cite{Ledda+2005,Eilertsen+2013}, image compression~\cite{Artusi+2016}, and single-image HDR~\cite{hanji2022comparison}. To the best of our knowledge, we present the first study on perceptual assessment of NVS methods and hope that our study will similarly inspire improvements that better meet the needs of human users.

\subsection{NVS Datasets}
\ac{NVS} methods are typically evaluated using synthetic and real-world datasets with sparse views~\cite{llff, NeX, deepview, NSVF, blendedmvs, dtu}. The NeRF synthetic dataset \cite{original_nerf} consists of 8 inwards facing scenes rendered with blender \cite{blender}, each containing 200 test images rendered at viewpoints located spirally at the upper hemisphere around the object. The \dsllff{ dataset \cite{llff} is a forward-facing dataset of real scenes, but with very sparse test views. The DTU \cite{dtu} Stereo dataset is also widely used to evaluate novel view synthesis performance, but its captured views are too sparse to create a continuous video. Recently, De Luigi et al. \cite{scannerf} set up a resource-efficient system to capture 360-degree dense views of various objects, but only for simple objects in a controlled lab environment and without video references. RealEstate10K \cite{zhou2018stereo} only contains a fly-through style video per scene, where objects are only visible in a few frames and from limited angles, making it inaccessible to split into distinct train and test frames. Tanks and Temples \cite{knapitsch2017tanks} is designed for large-scale scenes and can introduce unfairness in comparison. The presence of moving people/objects in these datasets also makes them unsuitable for many NVS methods. In contrast, our dataset is the first forward-facing dataset that captures scenes with reference videos in both laboratory and fieldwork environments, with accurately calibrated poses for each video frame.

\section{\datasetName{} Dataset}
\label{sec:dataset}

\begin{figure} 
    \centering
    \includegraphics[width=1.0\columnwidth]{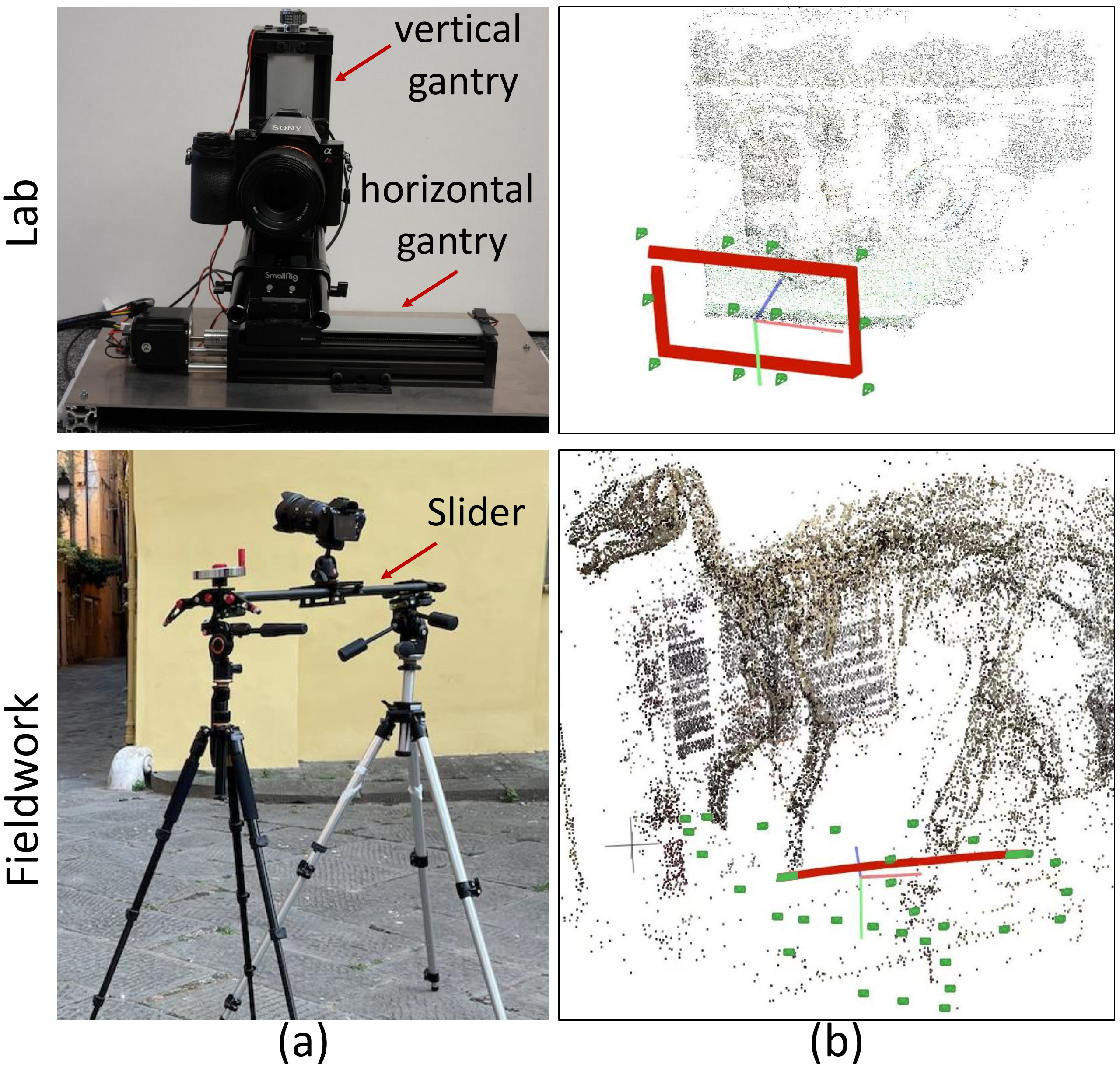}
    \caption{The camera rig (a) and camera poses (b) that are used to capture scenes for the \dslab{} (top) and \dsfieldwork{} (bottom) datasets. On the right, the green dots represent training camera poses and the red dots represent the poses of the reference video frames used for testing. }
    \label{lab-vis}
\end{figure}

To evaluate NVS methods on video rather than individual views, we collected two new datasets: \dslab{}, captured using a 2D gantry in a laboratory with controlled lighting and background; and \dsfieldwork{}, captured in-the-wild, consisting of both indoor and outdoor scenes. Both datasets were captured with Sony A7R\uppercase\expandafter{\romannumeral3}. Images of selected scenes from both datasets are shown in Figure~\ref{data-vis}.

\subsection{Lab Dataset}
\textbf{Capture Setup} The \dslab{} dataset was captured in our laboratory using a 2D gantry (upper-left of Figure~\ref{lab-vis}), which allowed horizontal and vertical movement of a camera. To minimize the amount of noise and avoid saturated pixels, we captured each view with a RAW image stack consisting of 2 exposures at constant ISO. The RAW image stacks were merged into an HDR image using an estimator that accounts for the photon noise \cite{hanji2020noise}. All images were color-corrected using a reference white point and cropped to $4032\times3024$\,px. 
To map linear images to display-referred units, gamma was applied ($\gamma=2.2$).

A sparse set of training views were taken on a uniform 
grid, shown as green dots in upper-right of Figure \ref{lab-vis}. The training views cover a horizontal range of 100\,mm and a vertical range of 80\,mm. 
The reference videos consisted of 300 to 500 frames, captured in a rectangular camera motion, as shown by red dots in the figure. The camera traveled about 0.6\,mm between each frame. Since we only consider a view interpolation task (no extrapolation), the reference frames were positioned within the range of the training views. 

\textbf{Scenes} The lab scenes were placed inside a box of $30\,\text{cm}\times41.5\,\text{cm}\times38\,\text{cm}$ for capturing.
As illustrated in the first row of Figure~\ref{data-vis}, they were designed to cover a wide range of objects with various materials, including glass, metal, wood, ceramic, and plastics. The layout of the objects was selected to introduce occlusions and to offer a good range of depth, which would fit within the depth-of-field of the camera. The dataset contains challenging view-dependent effects, such as diffraction on the surface of a CD-ROM, specular reflections from metallic and ceramic surfaces, and transparency of the glass. Six scenes were captured in this dataset. 

\textbf{Pose Estimation} For accurate pose estimation, 4 sets of 4 AprilTag markers were placed in each corner of the scene. The camera positions were selected to ensure that all markers were visible in each view and the images were later cropped to remove the markers. By detecting the position of AprilTags \cite{olson2011apriltag}, we obtained camera poses with standard camera calibration methods \cite{zhang2000flexible}. According to our pose estimation results, we got an acceptable mean re-projection error
of 0.2174 px across all scenes.
\subsection{Fieldwork Dataset} 
Our in-the-wild \dsfieldwork{} dataset was captured in both outdoor city areas and indoor rooms of a public museum\footnote{The location is not revealed for anonymity.}. Typically, such scenes are challenging due to complex backgrounds, occlusions, and uncontrolled illuminations.  

 
\textbf{Capture Setup} 
Different from the \dslab{} dataset, we captured video sequences instead of individual images for the \dsfieldwork{} scenes. 
The video sequences were captured with resolution  1920$\times$1080\,px and framerate 30\,fps. To reduce camera shake, we used either a DJI RS3 gimbal or a 90\,cm manual slider, which was fixed on two tripods, see lower-left of Figure~\ref{lab-vis}. 
For each scene, we captured several video sequences with different trajectories. One of these sequences, whose trajectory is well within the bounds of the scene, is selected as the test sequence. The bottom-right of Figure~\ref{lab-vis} shows the test sequence of one scene from the \dsfieldwork{} dataset (red dots). Images for training are sampled from the remaining videos (green dots). We also moved the first and last 15 frames from the test video sequence to the training set to ensure that the test views can always be interpolated from training views. In total, we have around 120 frames reserved as test views.

\textbf{Scenes} The second row of Figure~\ref{data-vis} shows selected examples of the captured scenes, which cover both indoor and outdoor scenarios with a high variability of materials including wood, marble, window glasses, metals, etc. and complex geometries such as a whale skeleton, posing challenging scenarios for NVS methods. Nine scenes were captured in this dataset.

\begin{figure*}[tb]
    \centering
    \includegraphics[width=\textwidth]{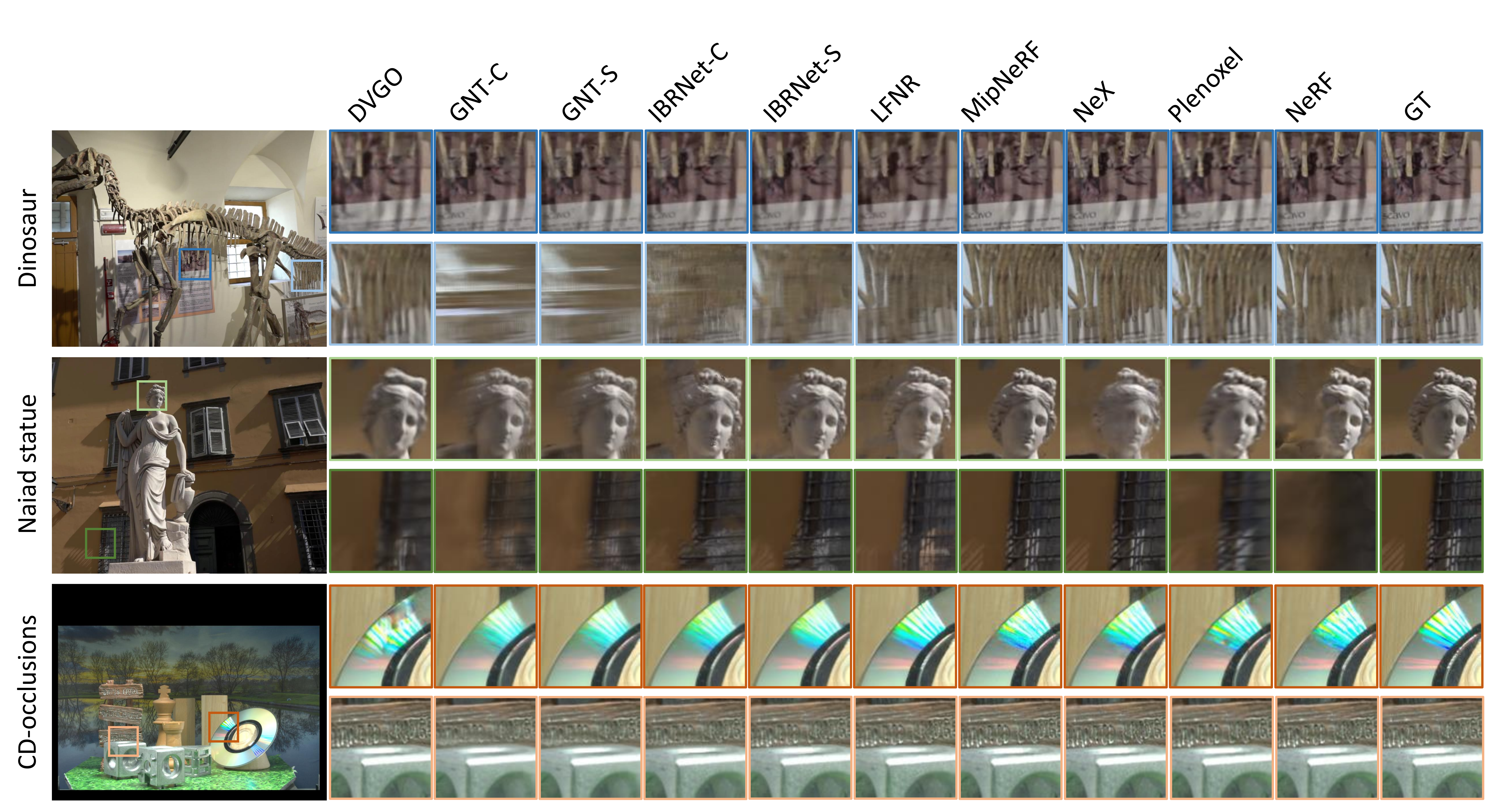}
    \caption{Examples of reconstructions by various NVS methods on selected scenes from \dsfieldwork{} dataset (first two rows) and \dslab{} dataset (third row). We only show three scenes due to limited space, please refer to the supplementary for more visual results.}
    \label{render_results}
\end{figure*}

\textbf{Pose Estimation} 
We employed COLMAP~\cite{schoenberger2016sfm} to perform joint calibration of camera poses for both the training and testing frames, so that all the calibrated poses share the same scale with a consistent coordinate system. We used the ``OPENCV" camera model, which supports separate x and y focal lengths as well as radial and tangential distortions. We also used COLMAP to undistort the captured images after pose estimation. Our reconstructed camera parameters have a mean reprojection error of 0.5327\,px across all scenes.

\subsection{Evaluated NVS Methods} \label{SEC:nerf-methods}

We tested ten representative NVS methods (including two generalizable NeRF variants) that encompass a diverse range of models, which feature both explicit and implicit geometric representations, distinct rendering modelings, as well as generalizable and per-scene optimization strategies. 
NeRF~\cite{original_nerf} is a neural volumetric representation that excels in image-based scene reconstruction and novel view synthesis.
Mip-NeRF~\cite{mipnerf} builds upon NeRF and provides a multiscale representation for anti-aliased view synthesis. 
DVGO~\cite{dirctvoxgo} and Plenoxels~\cite{plenoxel} use hybrid representations to achieve fast training and rendering.
NeX~\cite{NeX} utilizes multi-plane images and trainable basis functions, which is intended for rendering view-dependent effects in forward-facing scenes.
LFNR~\cite{lfnr} operates on a light field representation and uses an epipolar constraint to guide the rendering process. IBRNet~\cite{ibrnet} and GNT~\cite{gnt} are both generalizable NeRF models. IBRNet aggregates nearby source views to estimate radiance and density and GNT extends this idea by proposing a unified transformer-based architecture that replaces both multi-view feature aggregation and volume rendering. For IBRNet and GNT, we tested both their published cross-scene models (labeled as GNT-C and IBRNet-C) and also models fine-tuned on each scene (labeled as GNT-S and IBRNet-S).

We use these methods to reconstruct videos of scenes from both our collected datasets, \dslab{} and \dsfieldwork{}, as well as from the popular forward-facing LLFF~\cite{llff} dataset. For a fair comparison between methods, we downscaled images from the \dslab{} dataset by a factor of 4  and cropped images from the \dsfieldwork{} dataset, so that they all have the same training image resolution of $1008\times756$\,px. as that in LLFF scene. In this way, we were able to adopt the same training setup (network architecture, training iterations, optimizer, etc.) on LLFF scenes proposed by the respective authors. Please refer to the supplementary materials for more details about the training setup.

\section{Subjective Evaluation}
\label{sec:iq-experiment}
To attain precise subjective quality scores of the videos synthesized by the aforementioned NVS methods, we conducted a controlled quality assessment experiment with human participants. We relied on a pairwise comparison experiment, as it has been shown to be more accurate and robust than direct rating methods \cite{pair_comparison}. 
\subsection{Experimental Procedure}
\label{sec:experimental_procedure}
In each trial of our experiment, a participant was shown a pair of videos side-by-side on the same display and was instructed to pick the video of higher quality --- ``better resembles a natural scene and contains fewer distortions" (exact wording on the briefing form).
To reduce the number of comparisons and maximize the information gained from each trial, we used ASAP \cite{mikhailiuk2020active}, an active sampling method. 
Participants could press the space bar to view the reference video of the displayed scene (except for \dsllff{} dataset as reference videos were not available). The reference videos were included as one of the compared conditions. %
\subsection{Display and Videos}
\label{sec:STIMULI:VIDEO}
Our videos were displayed on a 27" Eizo ColorEdge CS2740 4K monitor, which was colorimetrically calibrated to reproduce BT.709 color space with a peak luminance of 200\cdms. The average viewing distance was 70\,cm, restricted by a table in front of the display.

We used 14 scenes from our two datasets and 8 scenes from the \dsllff{} dataset. For the scenes in our dataset, videos were synthesized on the same views as in ground-truth video frames. As \dsllff{} dataset does not have reference videos, we combined 120 frames rendered in a spiral trajectory around the mean pose (similar to other NVS methods). All the video frames were cropped to a resolution of $960{\times}756$\,px. and then up-scaled to $1920{\times}1512$\,px. (bilinear filter) so that two videos could be shown side by side on our 4K monitor. The upscaling was necessary to obtain a more realistic and effective resolution of 40 pixels per degree with respect to the original video resolution. Please note that when we evaluate image/video metrics in Section \ref{sec:metrics}, all the computation is also done on up-scaled images/videos to ensure fairness. Each video was between 3 to 15 seconds long, with a framerate of 30 fps. In total, each participant assessed the quality of 22 scenes reconstructed by 10 NVS methods as well as 14 reference videos.
\subsection{Participants}
We invited  39 volunteers (20 males and 19 females)
with normal color vision (confirmed by running the \textit{Ishihara Test}). Each participant completed 4--5 full batches of comparisons scheduled by ASAP \cite{mikhailiuk2020active}. 
 The experiment was authorized by an external institutional review board and the participants were rewarded for their participation. 

 \subsection{Subjective Score Scaling and Calculation}
\label{jod_scale}
We scaled the results of the pairwise comparison and expressed the subjective evaluation score in Just-Objectionable-Difference (JOD) units using the Thurstone Case V observer model \cite{pwcmp}. A difference of 1\,JOD unit means that 75\% participants preferred one method over another. The model assumes that participants made their selections by assigning a single quality value to each video and approximates this quality by a normally distributed random variable with the same inter- and intra-observer variance.
\section{Perceptual Benchmark Results}
\label{benchmark}
\begin{figure}[t]
\centering
    \includegraphics[width=\columnwidth]{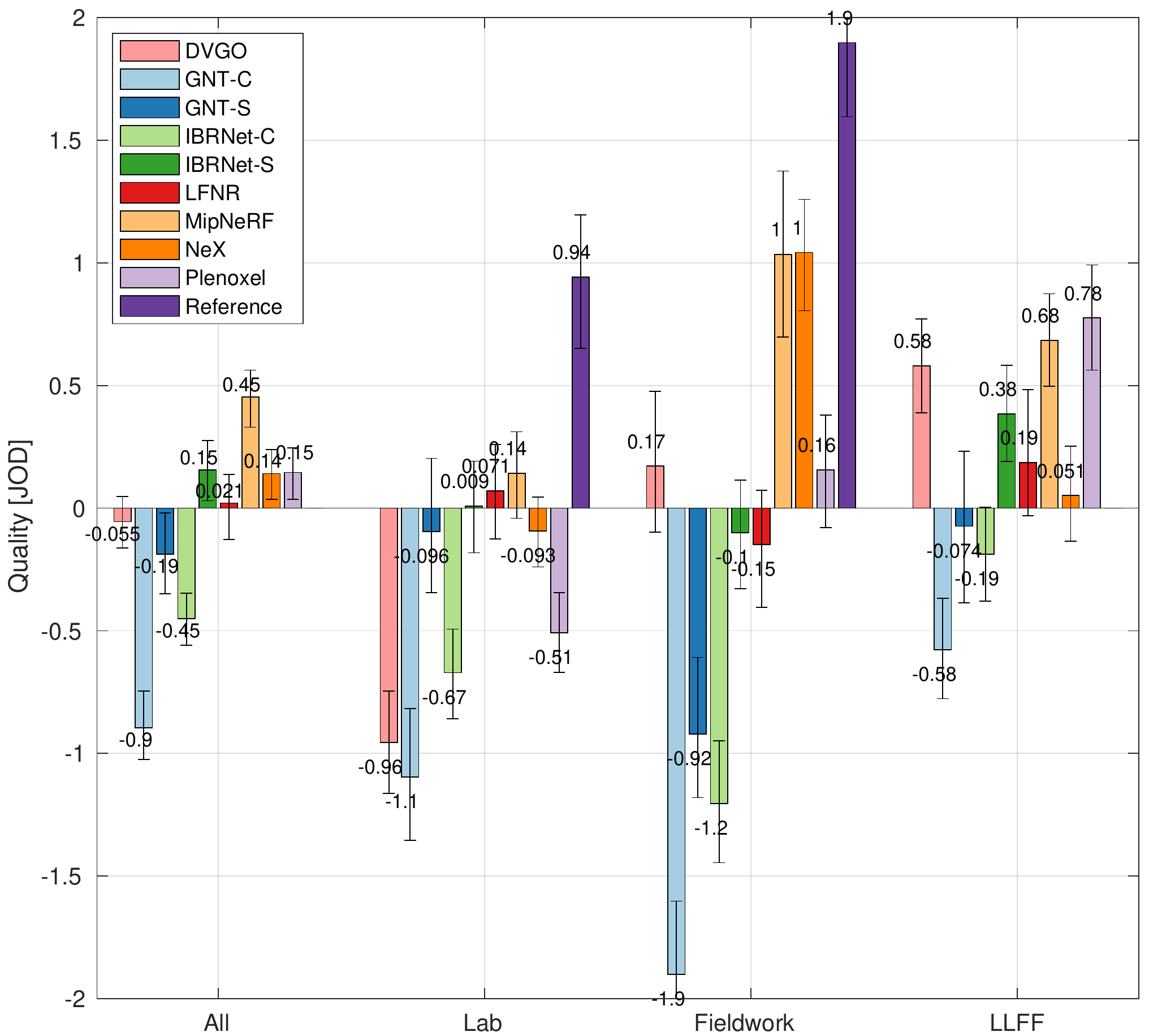}
   \caption{Perceptual preferences for NVS methods. The bars indicate preference in JOD units, relative to the original NeRF method~\cite{original_nerf}, which is at 0\,JOD. Negative values indicate that, on average, the method produced less preferable results than NeRF. The error bars show 95\% confidence intervals.} 
\label{fig:perceptual_score}
\end{figure}

\begin{table}
\caption{The list of evaluated objective metrics. 
}
\resizebox{\columnwidth}{!}{
\begin{tabular}{|c|c|c|l|}
\hline
\multirow{2}{*}{\textbf{Metric}} & \textbf{Reference} & \textbf{Video}       & \multirow{2}{*}{\textbf{Details}}   \\
                                 & \textbf{required}  & \textbf{metric}    &                                     \\ \hline \hline
\multirow{2}{*}{PSNR} & \multirow{2}{*}{\cmark} & \multirow{2}{*}{\xmark} & Widely used ratio to measure noise \\
& & & relative to the signal in log units \\ \hline
\multirow{2}{*}{PSNR-L} & \multirow{2}{*}{\cmark} & \multirow{2}{*}{\xmark} & PSNR computed on image luma values   \\
 & & &  \\ \hline
SSIM & \multirow{2}{*}{\cmark} & \multirow{2}{*}{\xmark} & Popular quality measure that perceives  \\
\cite{wang2004ssim} & & & structural similarity \\ \hline
MS-SSIM & \multirow{2}{*}{\cmark} & \multirow{2}{*}{\xmark} & \multirow{2}{*}{Multi-scale version of SSIM} \\
\cite{wang2003msssim} & & & \\ \hline
VIF & \multirow{2}{*}{\cmark} & \multirow{2}{*}{\xmark} & \ac{NSS} models \\
\cite{sheikh2006vif} & & & on information-theoretic setting \\ \hline
FSIM & \multirow{2}{*}{\cmark} & \multirow{2}{*}{\xmark} & Low-level image feature similarity \\
\cite{fsim} & & & based on the human visual system \\ \hline
LPIPS-VGG & \multirow{2}{*}{\cmark} & \multirow{2}{*}{\xmark} & Perceptual similarity metric based on \\
\cite{lpips} & & & deep network of VGG model \\ \hline
LPIPS-ALEX & \multirow{2}{*}{\cmark} & \multirow{2}{*}{\xmark} & Perceptual similarity metric based on \\
\cite{lpips} & & & deep network of AlexNet model\\ \hline
DISTS & \multirow{2}{*}{\cmark} & \multirow{2}{*}{\xmark} & Unify texture and structure similarity   \\
\cite{dists} & & &  with deep network\\ \hline
HDR-VDP-3 & \multirow{2}{*}{\cmark} & \multirow{2}{*}{\xmark} & Low-level vision model  on \\
\cite{Mantiuk2011} & & & HDR images \\ \hline
FLIP & \multirow{2}{*}{\cmark} & \multirow{2}{*}{\xmark} & Metric that considers HVS, viewing  \\
\cite{flip} & & & distance and monitor conditions\\ \hline
FovVideoVDP & \multirow{2}{*}{\cmark} & \multirow{2}{*}{\cmark} & Spatial-temporal metric
that accounts \\
\cite{fovvdp} & & & for foveation effect\\ \hline
STRRED & \multirow{2}{*}{\cmark} & \multirow{2}{*}{\cmark} & Hybrid metric measures temporal \\
\cite{strred} & & &  motion and spatial difference\\ \hline
VMAF & \multirow{2}{*}{\cmark} & \multirow{2}{*}{\cmark} & Support Vector Machine combination \\
\cite{vmaf1} & & &  of multiple image and video metrics\\ \hline
HDR-VQM & \multirow{2}{*}{\cmark} & \multirow{2}{*}{\cmark} & Spatial-temporal metric that considers \\
\cite{hdrvqm} & & &   human eye fixation behavior\\ \hline
BRISQUE & \multirow{2}{*}{\xmark} & \multirow{2}{*}{\xmark} & Support vector regression trained \\
\cite{brisque} & & & on IQA dataset \\ \hline
NIQE & \multirow{2}{*}{\xmark} & \multirow{2}{*}{\xmark} & Distance between \ac{NSS}-based \\
\cite{niqe} & & & features to those from a database \\ \hline
PIQE & \multirow{2}{*}{\xmark} & \multirow{2}{*}{\xmark} & Averaged block-wise distortion \\
\cite{piqe} & & & estimation \\ \hline
\end{tabular}
}
    \label{tab:metrics}
    \vspace{-3mm}
\end{table}

\figref{perceptual_score} shows the perceptual preference for different methods averaged across our collected \dslab{} and \dsfieldwork{} datasets, as well as the \dsllff{} dataset. We report both per-dataset performance and the overall performance across all three datasets. To view results on the individual scenes, we refer to Figures 2--4 
in the supplementary. The baseline (0 JOD line) in \figref{perceptual_score} is the original NeRF model~\cite{original_nerf}, so positive JOD values indicate improvement and negative values indicate degradation in quality (on average) with respect to NeRF.

The results on both of our datasets show that despite the impressive performance of \ac{NVS} methods, their results can still be easily distinguished from the reference videos. There is about 0.85\,JOD difference between the best neural rendering methods and the reference; 0.94 \textit{vs.} 0.14 for \dslab{}, 1.9 \textit{vs.} 1 for \dsfieldwork{}. This indicates that the reference will be selected as better in 70\% of the cases across the population. 
On average, only five out of nine methods produced better results than the original NeRF. It is evident that existing generalizable models require further refinement, as an additional per-scene optimization step is needed to achieve desirable outcomes.  It is noteworthy that the discrepancies among the methods are more noticeable in the more  challenging \dsfieldwork{} dataset, which implies that a challenging dataset is essential to distinguish between methods.


Compared with other models, MipNeRF performs quite well in most scenes, particularly those with high-frequency geometric details (Figure~\ref{render_results}, statue's face and hair in \scene{Naiad statue}, bones and background poster in \scene{Dinosaur}, fence in \scene{CD-occlusions} \textit{etc.}).
In comparison, techniques that lack explicit volume rendering (e.g., LFNR, and GNT) and those with coarse geometric modeling (e.g., NeX) exhibit suboptimal performance in these situations. 
Nonetheless, LFNR and NeX do provide more natural outcomes for scenes with complex lighting and specular reflections such as \scene{Metal} and CD in \scene{CD-occlusions}; see Figure~\ref{render_results}. For certain \dsfieldwork{} scenes, techniques founded on multi-view epipolar geometry constraints, such as IBRNet and GNT, tend to fail and exhibit conspicuous artifacts (\scene{Dinosaur} in Figure~\ref{render_results}, \scene{Vespa} and \scene{Giraffe} shown in supplementary). This is due to the considerable distance between the test and source views, which renders the epipolar features inaccurate. For a more in-depth examination, we encourage the reader to review the quality results of individual scenes provided in the supplementary. 
\section{Assessing Quality Metrics for Neural View Synthesis} 
\label{sec:metrics}


\begin{figure}[t]
\centering
\includegraphics[width=0.98\linewidth]{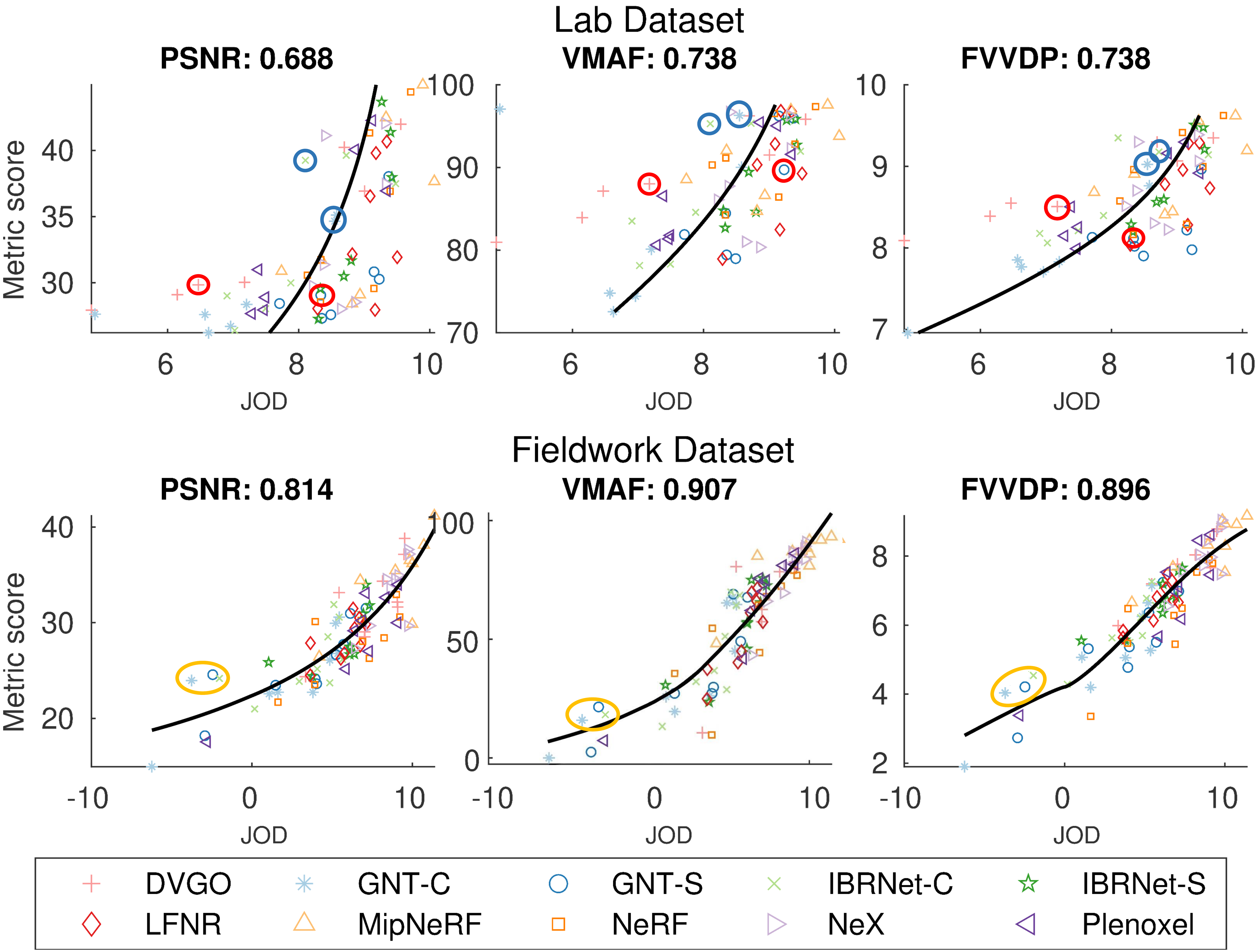}
\caption[]{Selected metric correlations for our \dslab{} (top row) and \dsfieldwork{} (bottom row) datasets. The black lines are obtained through fitting a logistic function which helps to detect outliers that affect correlations. \scene{Glossy animals} produced by DVGO and GNT-S are denoted in red circles in the first row. \scene{Metal} produced by IBRNet-C and GNT-C are denoted in blue circles. \scene{Vespa} produced by IBRNet-C, GNT-S, and GNT-S are denoted in orange circles in the second row.}
\label{fig:metric-scatter}
\vspace{-2mm}
\end{figure}

\begin{figure*}
    \centering 
    \includegraphics[width=0.95\linewidth]{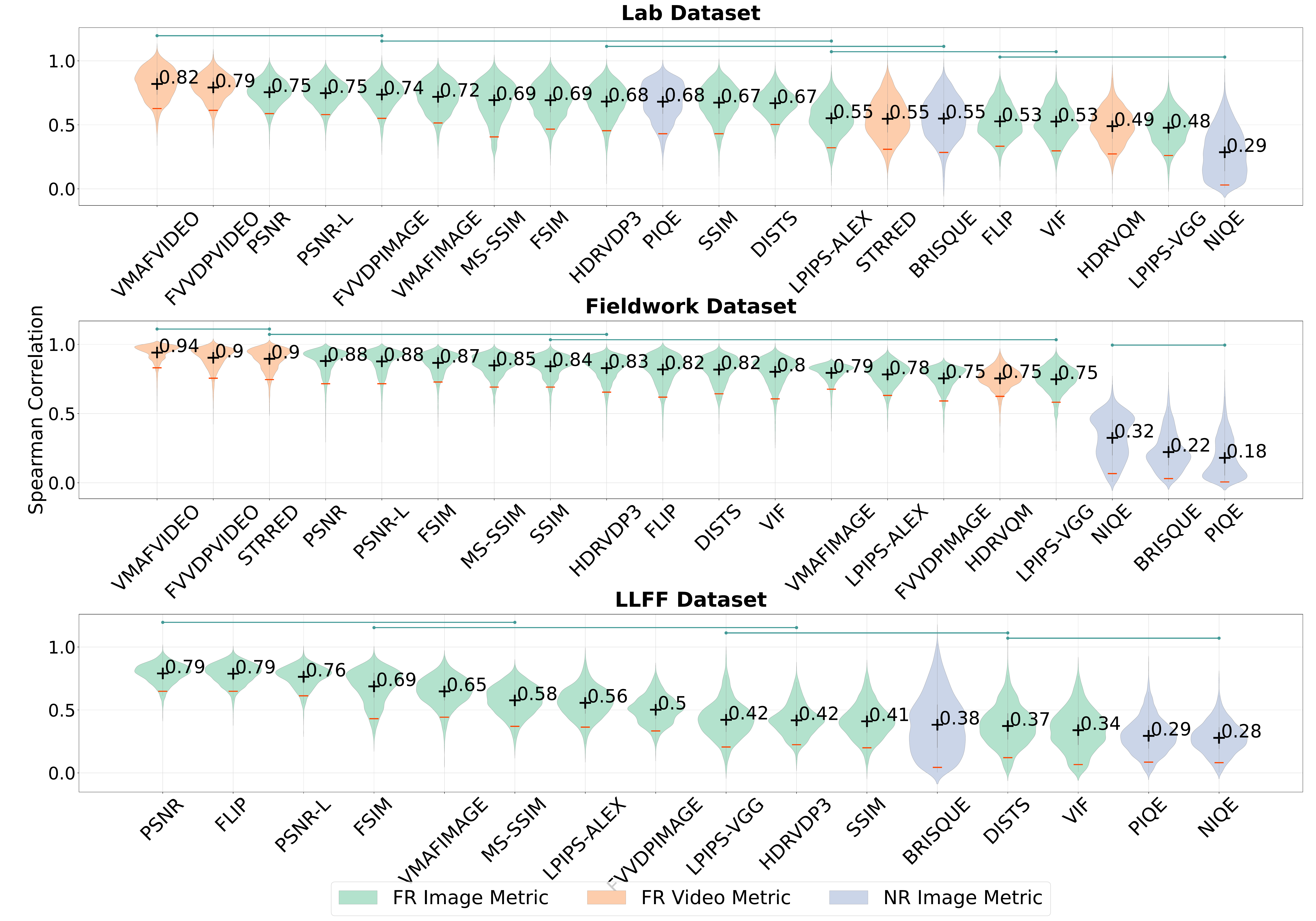}
    \caption{Bootstrapped distributions of correlation coefficients for all metrics
    computed on (a) \dslab{}, (b) \dsfieldwork{}, and (c) \dsllff{}. The ``+" in black denotes mean correlation, and ``{\color{red}-}" in red denotes the 5th percentile (an estimate of the bad-case performance). Full-reference video metrics are missing for \dsllff{} because this dataset has no ground-truth videos. The lines connect metrics where differences cannot be deemed statistically significant in a non-parametric test, with a $p$-value of 0.05.}
    \label{fig:correlations-all}
\end{figure*}

Our 
collected datasets with video references, together with perceptual quality results of reconstructed videos, allow us to test how well the existing image/video quality metrics can measure the perceived quality. 
We test a range of existing objective metrics, full-reference and non-reference, image and video metrics, as listed in \tableref{metrics}. We look into the widely used image similarity metrics such as PSNR and SSIM~\cite{ssim}, and also deep-learning related metrics, such as LPIPS~\cite{lpips} and DISTS~\cite{dists}. We test LPIPS using two different backbone models, VGG and AlexNet, as we noted they differ in their predictions. PSNR-L converts image RGB values into luma values before computing PSNR similarity. We include several video quality metrics, including FovVideoVDP (v1.2, labeled FVVDP) \cite{fovvdp}, VMAF (v0.6.1)\cite{vmaf0,vmaf1,vmaf2}, STRRED and HDRVQM. Given that FVVDP and VMAF are applicable to both videos and images, we assess their performance in two distinct contexts: one is directly evaluated on videos (denoted with a "video" suffix), while the other is evaluated on each video frame and average among them (denoted with an "image" suffix). The absence of a suffix implies the evaluation is conducted on video content. In contrast to its image-based counterpart, a direct evaluation of videos entails the consideration of temporal distortions between successive video frames. For instance, VMAF-video incorporates an additional measurement of temporal differences within the luminance component when compared to VMAF-image. We also evaluated on several blind or non-reference image metrics (BRISQUE~\cite{brisque}, PIQE~\cite{piqe} and NIQE~\cite{niqe}), that directly compute scores without comparing to the reference images. 
For metrics that require display parameters (e.g., HDRVDP-3, and FovVideoVDP), we matched the effective resolution of the videos in our experiment (40\,ppd, see \secref{STIMULI:VIDEO}). 

For our collected \dslab{} and \dsfieldwork{} datasets, which include reference videos, we computed the quality scores on the captured test video sequences. Since \dsllff{} dataset lacks reference videos, we computed the quality scores on the sparse test image set as done in NVS papers. Thus, it is inherently unfeasible to consider the temporal distortions within this dataset, which is likely to be inferior to testing on videos.

To test the reliability of popular metrics, we followed the standard
protocol used to evaluate quality metrics~\cite{hanji2022comparison,Ponomarenko2015}, and computed rank-order (Spearman) correlations between metric predictions and perceptual JOD values. 
Figure \ref{fig:metric-scatter} shows scatter plots and correlations of popular quality metrics w.r.t. subjective scores for the \dslab{} (top row) and \dsfieldwork{} (bottom row) datasets. Please refer to the supplementary for similar plots for other metrics. We use least square optimization to fit a logistic function between metric score and subjective JOD score, which helps us find the outliers that affect the correlations. 
 However, these point estimates of correlations conceal measurement noise due to: (a) the selection of scenes, (b) the subjective experiment results. Thus we cannot draw conclusions solely based on these correlations.
\subsection{Averaged Bootstrapped Correlations}\label{sec:bootstrap-correlations}
For each dataset and each NVS method, we averaged subjective scores and quality metric predictions across all scenes and then computed a single correlation per dataset per metric. This serves two purposes: (a) it mitigates the effects of measurement noise and improves the predictions of quality metrics as shown in previous works \cite{hanji2022comparison}, and (b) NVS methods are typically compared on averaged scores across scenes that reduce per-scene bias, making it more relevant for us.

When comparing quality metrics, it is essential to account for the variance in our data (subjective score variance and scene selection). We estimate the distribution of correlation values using bootstrapping~\cite{Mooney1993}:
we generated 2000 bootstrap samples for each estimated correlation by randomizing (sampling with replacement) both the participants and the selection of scenes. Within each bootstrap sample, we independently scaled the JOD values (following Section~\ref{jod_scale}). In this way, our bootstrapping simulates 2000 outcomes of the experiment to capture the variance we can expect due to measurement noise. To determine whether the differences between the metrics are statistically significant, we performed a non-parametric test at the $\alpha=0.05$ level by directly computing the distribution of the difference of bootstrapped samples. The results of that test are visualized as horizontal green lines in \figref{correlations-all}.

\subsection{Quality Metrics Performance}\label{sec:metric-quality}

The correlations between metrics scores and subjective scores are shown in \figref{correlations-all}. Below, we discuss the main observations that can be made based on that data.  



{\textbf{PSNR is more accurate than SSIM and LPIPS} NVS methods are typically evaluated using image quality metrics such as PSNR, SSIM, and LPIPS. The results in \figref{correlations-all} show that the simplest metric, PSNR, performed significantly better than more complex SSIM and LPIPS. NVS evaluation clearly does not benefit from the statistics extracted by SSIM or deep features extracted by LPIPS. Poor performance of SSIM has been noted before \cite{Ponomarenko2015,Lin2019}, but it is still a popular metric because of its simplicity. The poor performance of LPIPS could be attributed to its training data consisting of small image patches with specific distortion types (noise, blur, compression-related, etc.) that are unlike NVS artifacts. 
We did not observe a statistically significant performance difference between PSNR-L (computed on luma) and PSNR (computed on RGB).

\textbf{Importance of video reference dataset.} In Figure \ref{fig:correlations-all}, we observe that per-metric correlations are the lowest for the \dsllff{} dataset and the highest for \dsfieldwork{} dataset. The low correlations for \dsllff{} could be partially explained by the fact that while the subjective experiment measured video quality, the metrics could only be run on individual test views due to the lack of reference videos in LLFF. 

To further investigate the importance of video reference, we experimented with sequences in which the number of available frames varied. We recomputed metric scores on progressively denser subsets of the test video frames (ranging from 10\% to 100\% of frames). \figref{sparseset} illustrates the effect of increasing frames for representative image metrics. Note that the reported values are bootstrapped correlations (see \secref{bootstrap-correlations}) between metric predictions and subjective scores. The figure shows a gradual increase in correlation as we use more video frames. This observation highlights the limitations of using sparse image sets for assessing perceived quality, which degrades the predictions of image metrics. 

All the above results indicate that the current objective evaluation protocol using a sparse image set is inadequate for assessing the perceptual quality of NVS methods applied to video generation. This underscores the rationale behind the development of our new datasets, which incorporate reference videos for testing.

\textbf{Video metrics outperform image metrics.} For both datasets with video reference (\dslab{} and \dsfieldwork{}), video quality metrics VMAF-video and FVVDP-video demonstrate the strongest correlations with human perceptual assessments. As shown in \figref{sparseset}, the predictions of video metrics (shown as star markers) were more accurate than those of the same metrics run on video frames (shown as square markers). This compelling evidence suggests that the subjective assessment of NVS-generated video is significantly influenced by temporal distortions and highlights again the importance of video assessment for NVS evaluation. This also emphasizes the significance of using video metrics that take temporal artifacts into account.

\begin{figure}[t]
\centering
\includegraphics[width=\columnwidth]{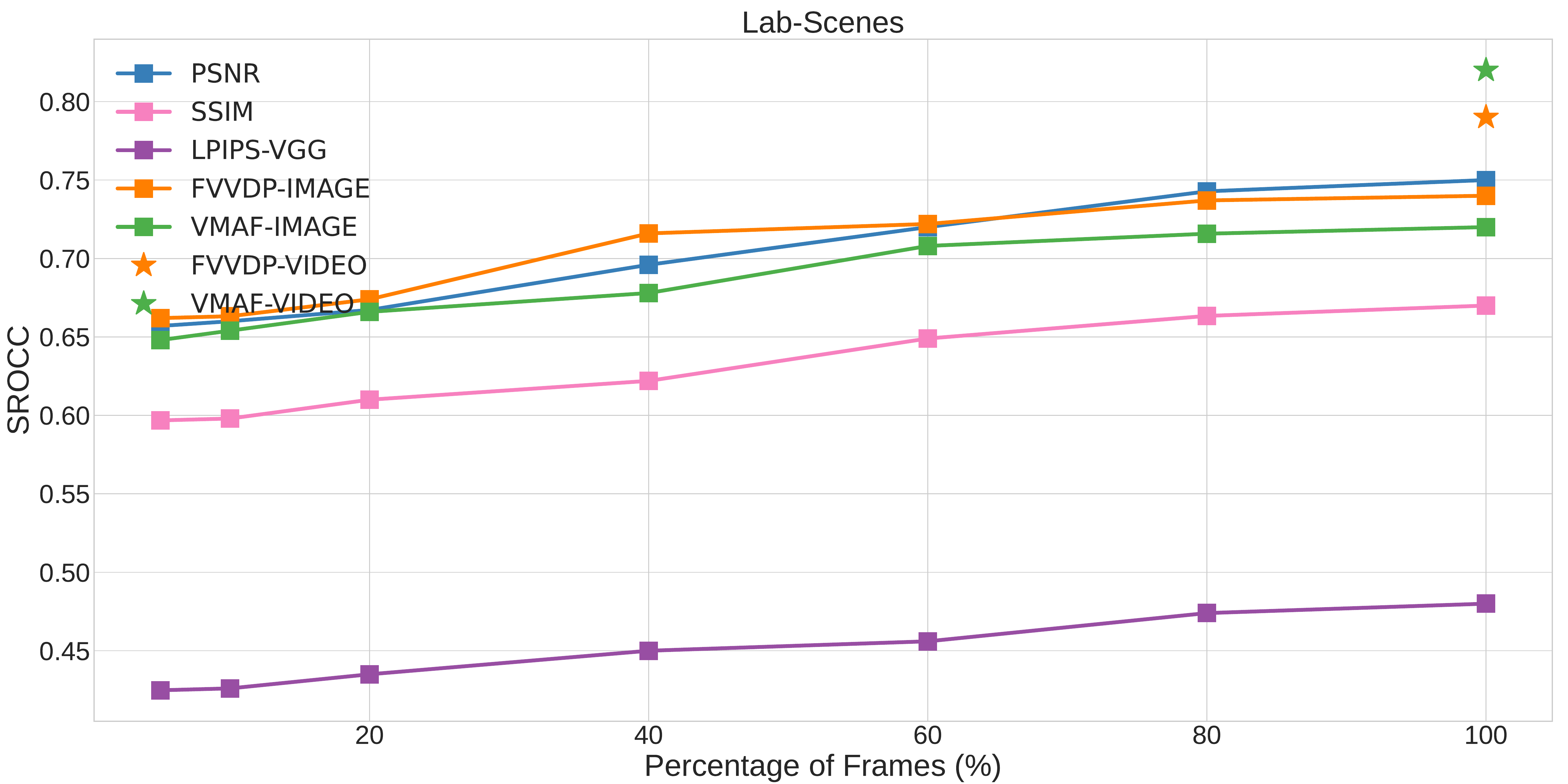}
\\
\vspace{3pt}
\includegraphics[width=\columnwidth]{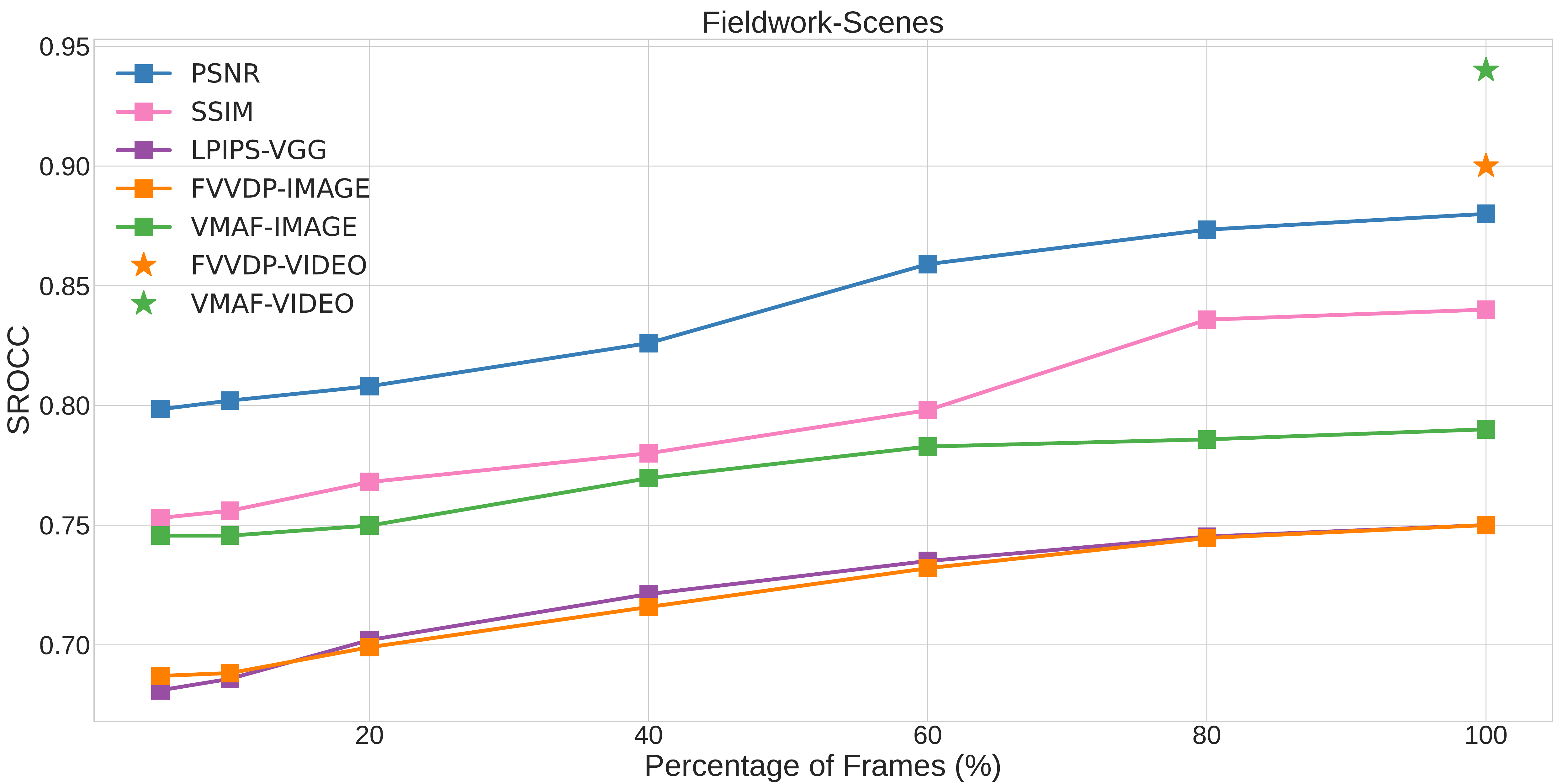}
\caption{The performance of selected image quality metrics (in terms of SROCC) as the function of the number of frames used for the quality prediction. The star symbols on the right denote the performance of video metrics run on all frames. The performance of image quality metrics improves as we supply more images. Still, the highest performance is achieved by video metrics (FVVFP-video, VMAF-video), which can detect temporal distortions.  
}
\label{fig:sparseset}
\end{figure}

\textbf{Importance of challenging datasets.} 
Similar to point estimates of correlations (Figure \ref{fig:metric-scatter}), the bootstrapped correlations are the highest for the \dsfieldwork{} dataset (\figref{correlations-all}). The simple explanation for this result is that the \dsfieldwork{} dataset was more challenging for NVS methods and resulted in larger, more objectionable artifacts, as can also be seen in the subjective results (\figref{perceptual_score}). Such large differences make it much easier for the quality metrics to differentiate between the methods. In fact, most full-reference quality metrics performed well on this dataset. The \dslab{} dataset, with its highly specular materials, was designed to pose a challenge to NVS methods. However, as it has a denser and more regular set of training views, most NVS methods performed well on those scenes, making it harder for the metrics to differentiate between the methods.

In summary, we recommend testing NVS methods on video sequences and using video quality metrics, such as VMAF and FVVDP. We still recommend using PSNR because of its simplicity, relatively good performance, and because it allows comparison with existing studies. We further recommend conducting testing of NVS methods on challenging datasets equipped with reference videos, such as our \dsfieldwork{} dataset.

\begin{figure}[t]
\centering
   \begin{subfigure}[b]{0.5\textwidth}
\includegraphics[width=0.95\textwidth]{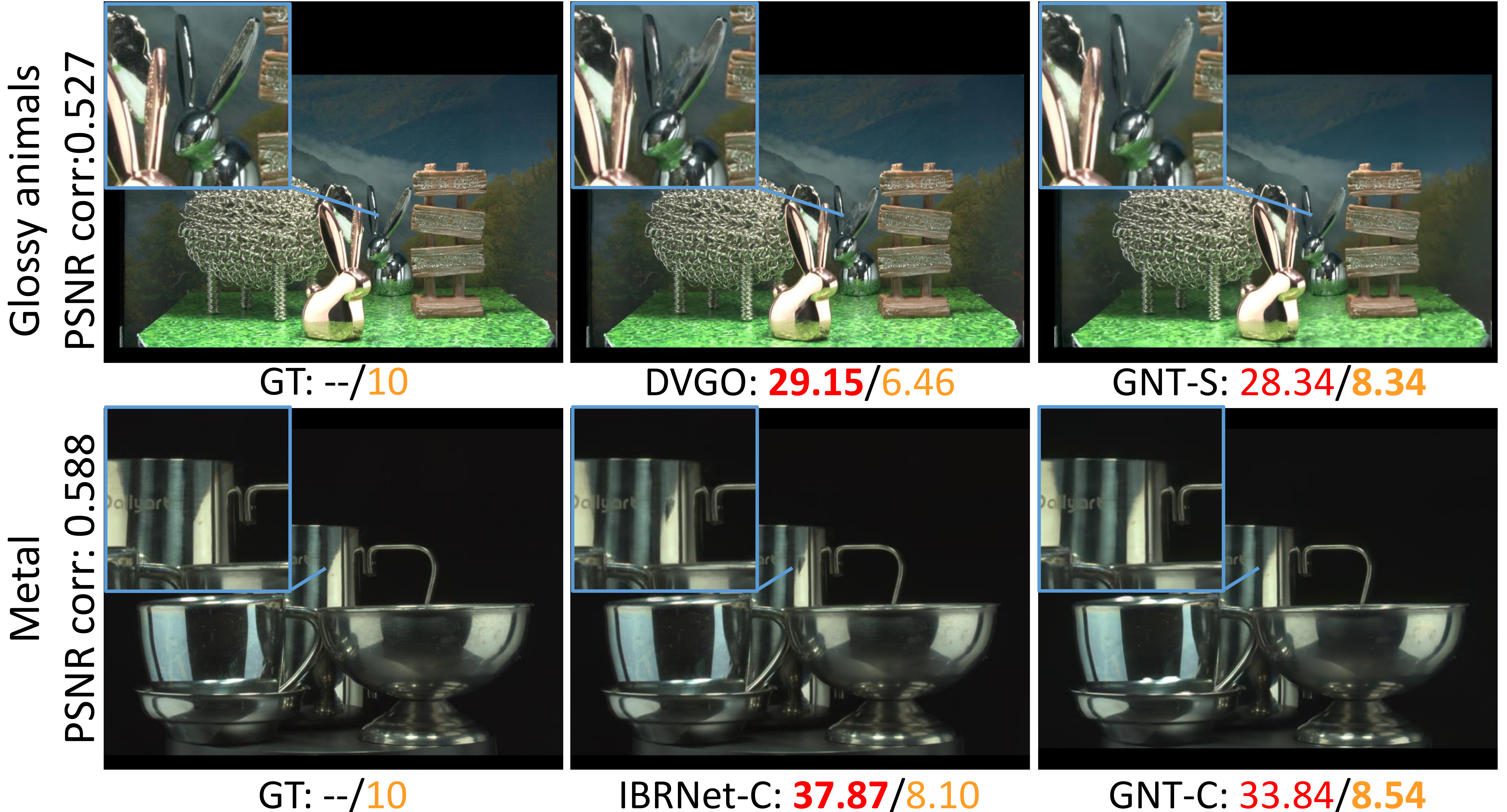}
   \caption{}
   \label{failurecase}
\end{subfigure}
\\
\begin{subfigure}[b]{0.5\textwidth}
\includegraphics[width=0.95\textwidth]{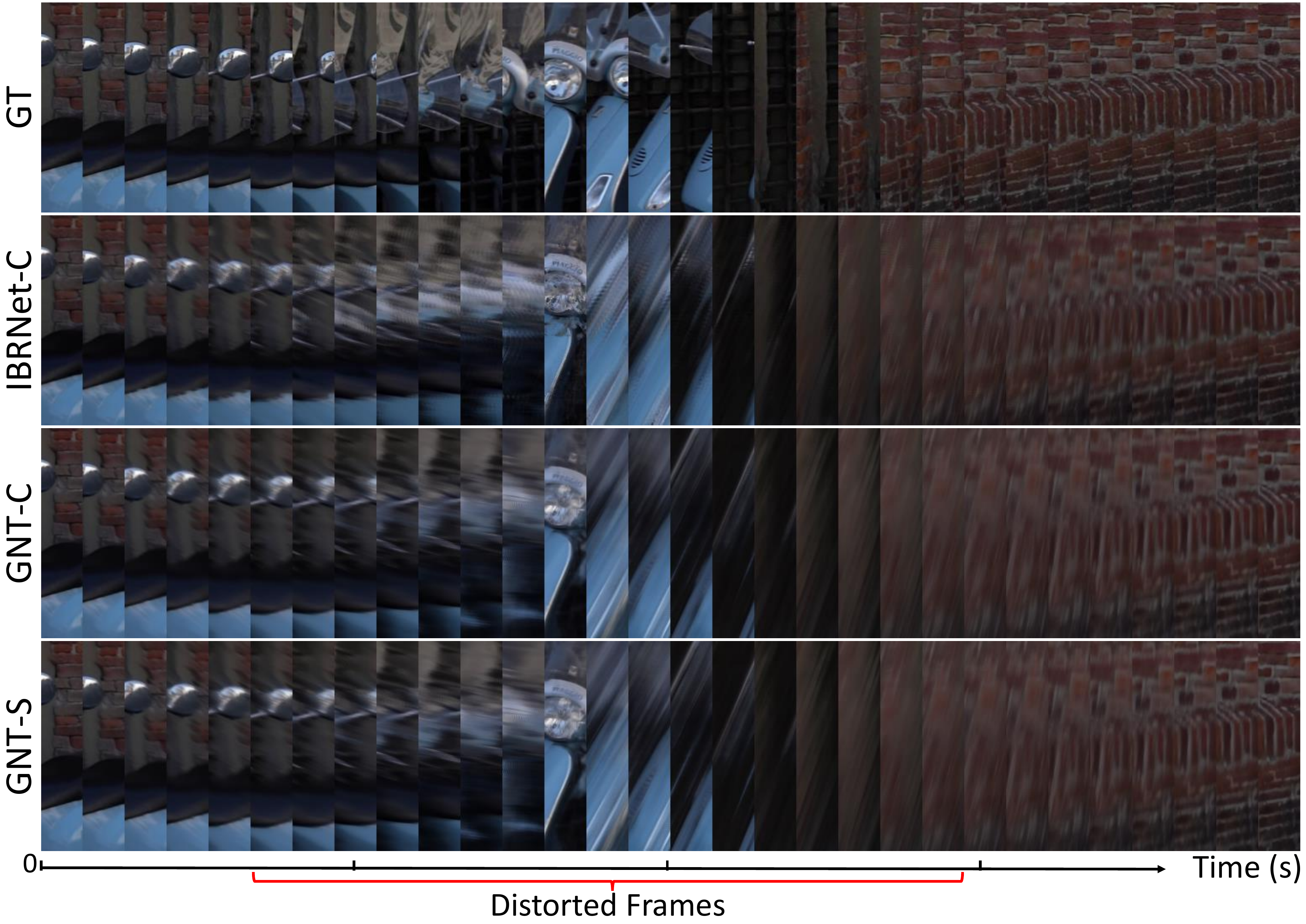}
   \caption{}
  \label{vespa}
\end{subfigure}
\caption{Representative scenes where PSNR fails to predict quality. (a) Scenes from \dslab{} dataset: The values under the plot show \textcolor{red}{PSNR} / \textcolor{YellowOrange}{subjective-JOD} values. The images in the middle column have higher PSNR values than those in the right column, but they contain obvious artifacts and are less preferred (compare JOD values) --- see the blurry artifact in \scene{Glossy animals} and unnatural local shading in \scene{Metal}. We show the Spearman correlation between PSNR and subjective score on the left of the GT image (\textit{i.e.} 0.527 on \scene{Glossy animals} and 0.588 on \scene{Metal}). (b) \scene{Vespa} from \dsfieldwork{} dataset produced by GNT-C, GNT-S, and IBRNet-C show a degradation in image quality from 1s to 4s with severe temporal distortions between consecutive frames. These distortions are markedly disfavored by study participants. However, PSNR fails to capture the distortions adequately and yields relatively favorable metric scores. Please note that all these artifacts are more noticeable in videos.}
\end{figure}

\subsection{Failure Cases of PSNR}
Although PSNR is an effective image metric for evaluating NVS methods with respect to the perceived subjective quality, it is still beneficial to investigate when PSNR can fail to reflect subjective preferences. To do so, we compute per-scene correlations between PSNR scores and bootstrapped perceptual JOD values,
and find scenes for which the metric results in poor correlations. 

On the \dslab{} dataset, we find that PSNR fails to accurately assess perceived quality on \scene{Glossy animals} and \scene{Metal} scenes. Particularly, for \scene{Glossy animals}, we observe that the NVS generated by DVGO exhibits a higher PSNR score compared to that produced by GNT-S, as depicted in Figure \ref{fig:metric-scatter}, however, the latter is more preferred by participants. This discrepancy is exemplified in Figure \ref{failurecase}, which presents a representative reconstruction of this scene. Notably, PSNR appears to lack sensitivity to the blurry artifacts introduced by NVS methods, as observed in the top row's middle column, specifically between the rabbit's ears. However, this artifact is easily noticeable (especially when shown in video) and significantly impacts human preferences, with a clear preference for images without such distortions (top row, right column in Figure~\ref{failurecase}). The inadequacy of PSNR is also underscored when examining the \scene{Metal} scene, notably in the NVS results produced by IBRNet-C and GNT-C, as illustrated in the second row of Figure \ref{failurecase}. In this scene, participants are highly sensitive to localized distortions, such as the unnatural local shading, as depicted in the bottom row's middle column in Figure \ref{failurecase}. These subtle yet crucial artifacts may not be effectively captured by PSNR due to its reliance on averaging pixel-level information across the entire image.

On the \dsfieldwork{} dataset, although PSNR effectively evaluates the perceived quality across most scenes, it exhibits limitations in adequately assessing the scene \scene{Vespa}, as can be regarded as an outlier illustrated in the orange circles in Figure~\ref{fig:metric-scatter}.  Specifically, the NVS results generated by GNT-C, GNT-S, and IBRNet-C, as depicted in Figure~\ref{vespa}, show a progressive degradation in image quality from 1s to 4s, accompanied by severe temporal distortions between consecutive frames. These observed distortions are markedly disfavored by study participants with low perceptual JOD value around -5. However, PSNR fails to capture the temporal distortions adequately, continuing to yield relatively favorable metric scores around 25. In contrast, both VMAF and FVVDP metrics prove to be effective in detecting and quantifying the presence of temporal artifacts (\scene{Vespa} produced by IBRNet-C/GNT-C/GNT-S has VMAF scores of 15.09/16.10/18.35, and FVVDP scores of 4.73/4.23/4.38), offering a better correlation with subjective scores.


\section{Conclusions}
The primary application of NVS methods is the interactive exploration of 3D scenes. Yet, those methods are typically tested on isolated views instead of videos, which could mimic such 3D exploration. In this work, we collected two new datasets with reference videos and used them to evaluate 8 representative NVS methods (and two variants) in a subjective quality assessment experiment. The results helped us to identify the strengths and weaknesses of tested NVS methods, but also to evaluate 18 image/video quality metrics. We found that (a) existing quality metrics struggle to differentiate between the NVS methods when they are tested on datasets with a dense set of training views; and (b) SSIM and LPIPS, which are two commonly used quality metrics, perform worse than PSNR when evaluating NVS methods. Our recommendation is to evaluate NVS methods on challenging datasets with sparsely sampled views and to use both PSNR and video metrics, such as VMAF and FovVideoVDP. 

\section*{Acknowledgements}
This work was supported by a UKRI Future Leaders Fellowship [grant number G104084].

{\small
\bibliographystyle{ieee_fullname}
\bibliography{egbib}
}

\appendix

\section{Supplementary Video}
We have included reference video clips of the scenes in both of our new datasets. However, we were unable to include the synthesized clips generated by the NVS methods due to file size restrictions on the supplementary materials. We will make them available on the project web page upon acceptance of the paper. 
\section{Per-scene Subjective Quality}

Due to space limitations, the main document contains subjective scores averaged across all scenes in each dataset (Figure~4 in the main document). Figures~\ref{fig:jod_lab},~\ref{fig:jod_fieldwork} and~\ref{fig:jod_llff} show the subjective results individually for each scene. These results show large variations across the scenes, but they also exhibit common trends:
\begin{itemize}
    \item The generalizable methods GNT and IBRNet perform poorly on all scenes in our new \dslab{} and \dsfieldwork{} datasets (worse than NeRF), but much better on the public \dsllff{} dataset. Per-scene fine-tuning (-S suffix) improves the predictions of both methods. 
    \item Similarly DVGO performs poorly on our new datasets, but much better on the \dsllff{} dataset.
    \item LFNR has rather uneven performance --- it is one of the best methods for some scenes (\dslab{}/CD-occlusions (I/E), \dslab{}/Glossy animals (I), \dsfieldwork{}/Naiad statue) but it fails in the others. 
    \item MipNeRF was one the most robust methods, performing typically better or on par with NeRF. In some of the scenes, it matched the quality of the reference (\dslab{}/Glass, \dsfieldwork{}/Leopards, \dsfieldwork{}/Giraffe, \dsfieldwork{}/Naiad statue, \dsfieldwork{}/Vespa). 
    \item Plenoxel performed well in most scenes in \dsllff{} dataset (except Room) but was generally worse than NeRF when tested on the \dslab{} dataset. Its performance varied from scenes to scene in the \dsfieldwork{} dataset, with a few fail cases (Dinosaur and Whale) but also better-than-NeRF performance (Leopards, Naiad statue, Vespa).
\end{itemize}
\begin{figure*}[t]
\begin{center}
    \includegraphics[width=0.8\textwidth]{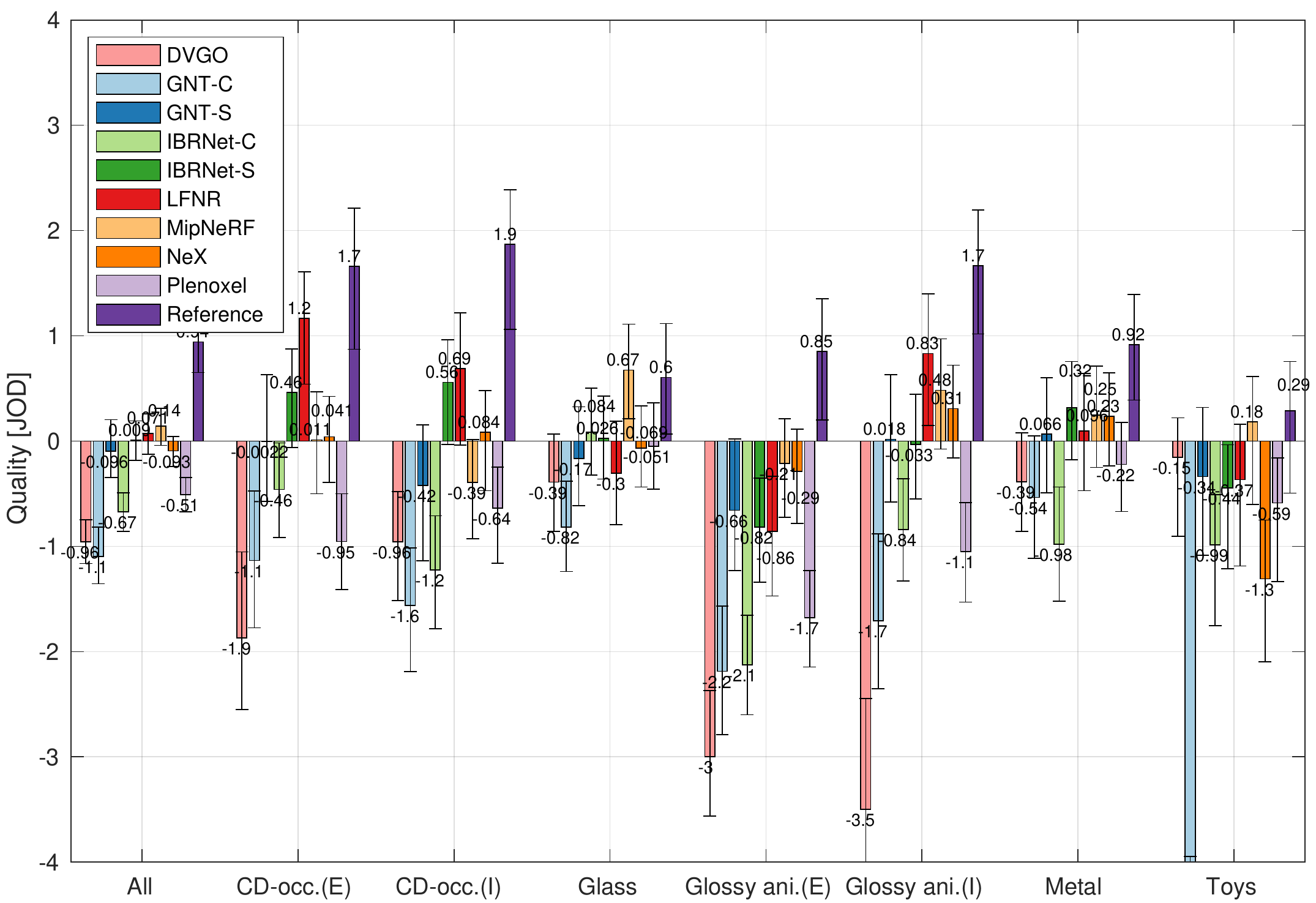}
\end{center}
   \caption{Perceptual preference of different NeRF methods on the \dslab{} dataset. The notation is the same as in Figure~4 in the main paper. The scenes with (I) in the label used the novel view selection that required only interpolation of the views, while the scenes with (E) required the views to be extrapolated.}
\label{fig:jod_lab}
\end{figure*}

\begin{figure*}[t]
\begin{center}
    \includegraphics[width=0.8\textwidth]{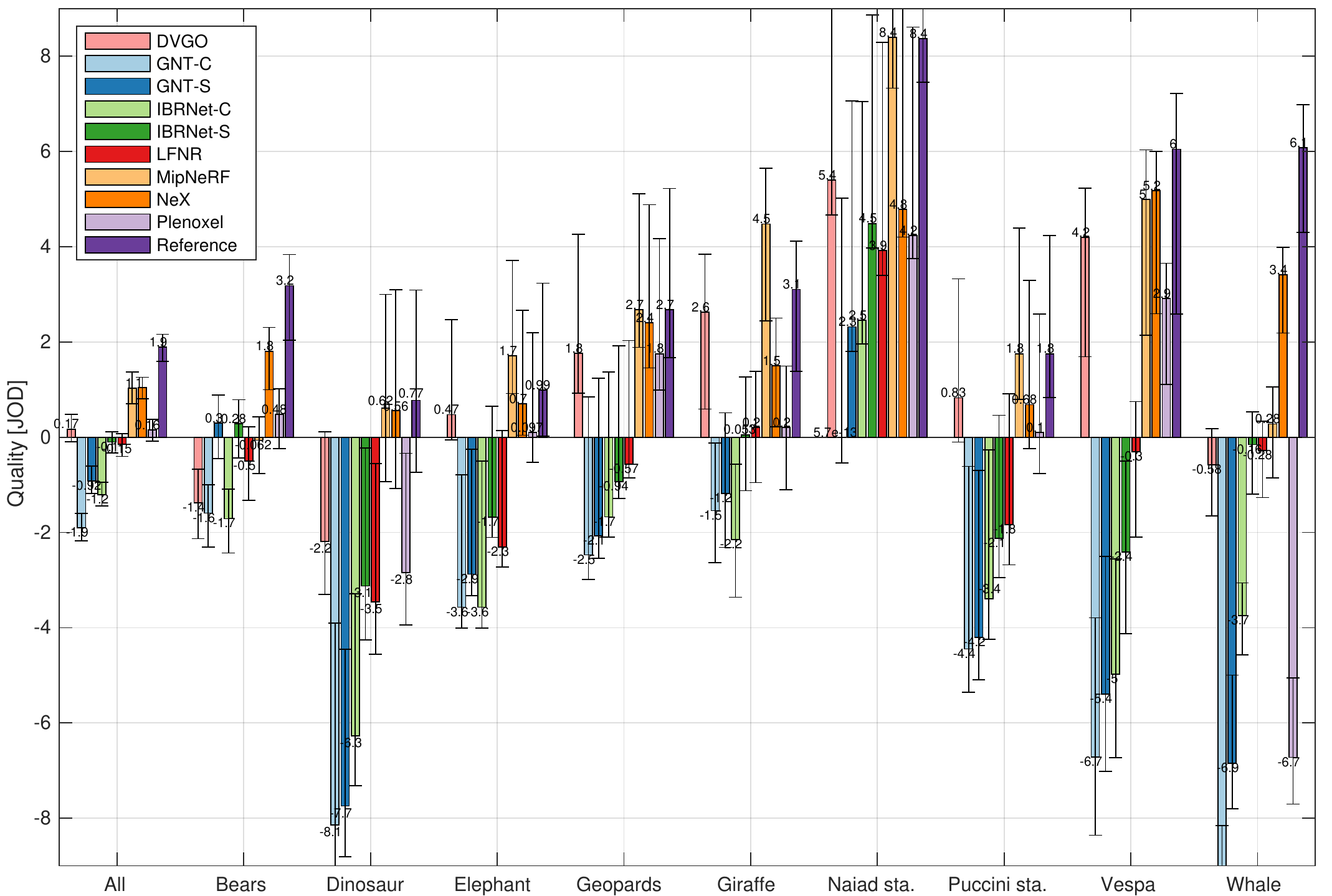}
\end{center}
   \caption{Perceptual preference of NeRF methods on the \dsfieldwork{} dataset.}
\label{fig:jod_fieldwork}
\end{figure*}

\begin{figure*}[t]
\begin{center}
    \includegraphics[width=0.8\textwidth]{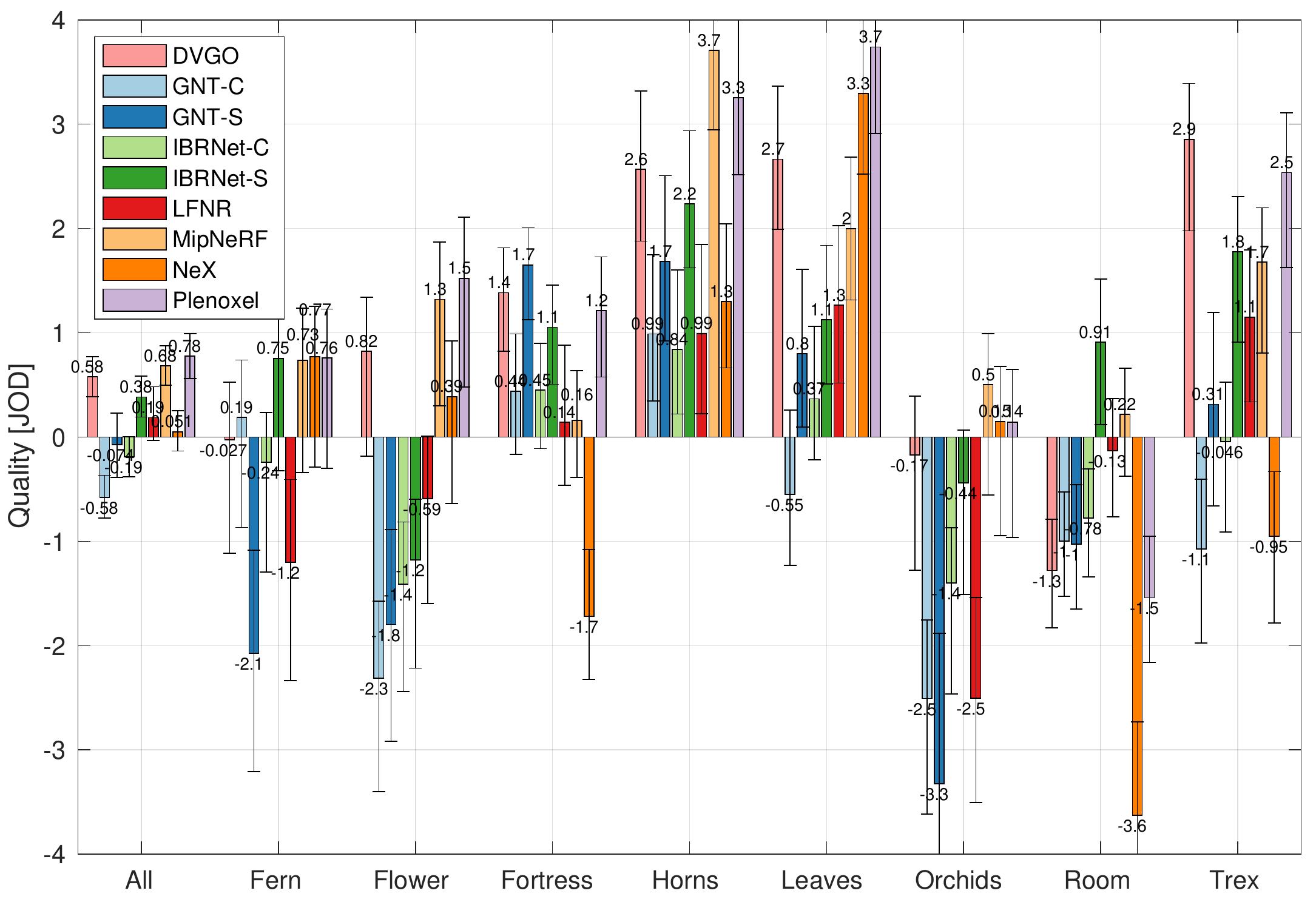}
\end{center}
   \caption{Perceptual preference of NeRF methods on the \dsllff{} dataset.}
\label{fig:jod_llff}
\end{figure*}

\section{Metric Performance: PLCC and RMSE}
Apart from Spearman Rank Order correlation, we also compute the boostrapped distribution of Pearson Linear Correlation Coefficient (PLCC) and Root Mean Squared Error between the image metrics score and perceptual quality score on each dataset. The results are shown in Figures~\ref{fig:pearson} and~\ref{fig:rmse}. With a few exceptions, the trends shown in those plots are similar to those shown for SROCC in Figure~6 of the main paper. The difference worth noting is that while the correlations (PLCC and SROCC) are much higher for the \dsfieldwork{} than for the \dslab{} dataset (indicating good metric performance), the opposite trend is shown by the RMSE. The RMSE values are on average smaller for the \dslab{} dataset, suggesting higher metric accuracy. It must be noted, however, that the range of subjective scores is much larger for the \dsfieldwork{} dataset (refer to the scatter plots in \figref{scatter_lab} and \figref{scatter_outdoor}). The difference in the RMSE numbers is most likely due to very different magnitudes of distortions in each dataset. If the goal of a metric is to differentiate between NVS methods, the correlation coefficients are better indicators of metric performance. 

\begin{figure*}[t]
\begin{center}
    \includegraphics[width=0.8\textwidth,height=0.5\textwidth]{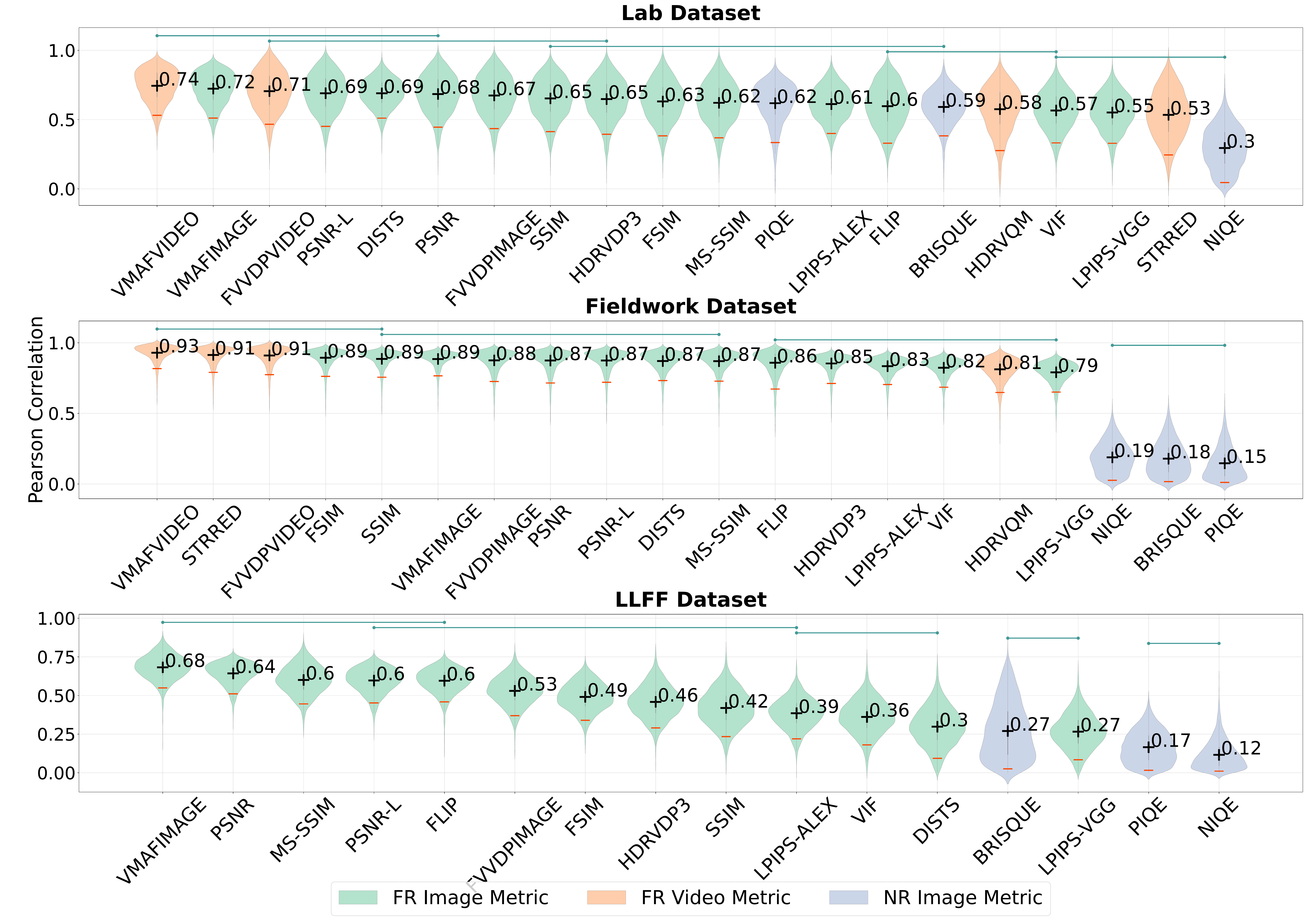}
\end{center}
   \caption{Bootstrapped distributions of Pearson Linear Correlation Coefficients (PLCC) for all metrics, reported separately for each dataset. The higher the number, the better is metric's performance. The notation is the same as for Figure~6 in the main paper.}
\label{fig:pearson}
\end{figure*}

\begin{figure*}[t]
\begin{center}
    \includegraphics[width=0.8\textwidth,height=0.5\textwidth]{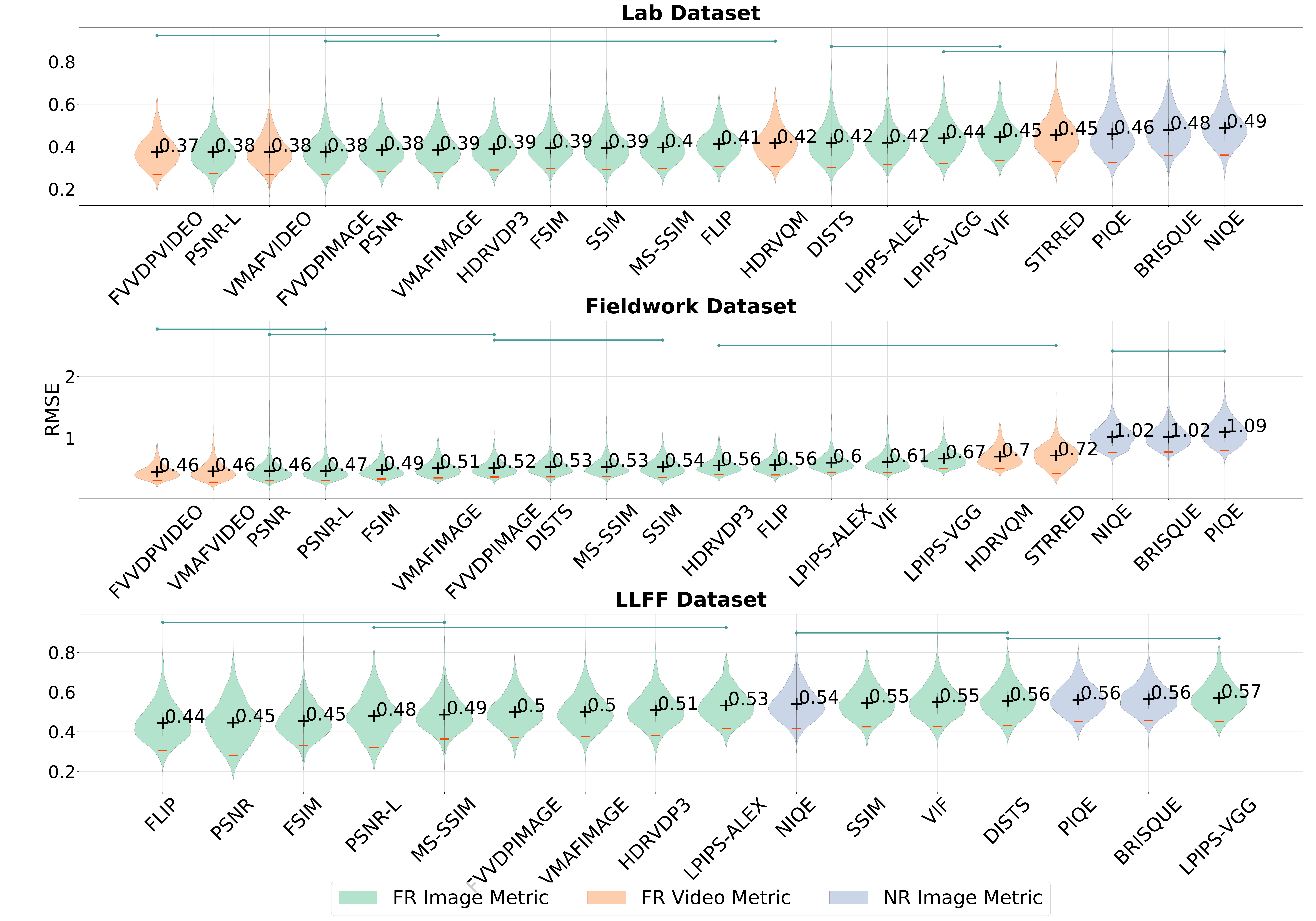}
\end{center}
   \caption{Bootstrapped distributions of Room Mean Squared Errors (RMSE) for all metrics computed separately for each dataset. The lower the number, the better is metric's performance. The notation is the same as for Figure~6 in the main paper.}
\label{fig:rmse}
\end{figure*}

\section{Metric Prediction Scatter Plots}
Metric predictions for individual scenes are compared with subjective scores in scatter plots in Figures~\ref{fig:scatter_lab}, ~\ref{fig:scatter_outdoor} and ~\ref{fig:scatter_llff}. When metric predictions are accurate, the scatter plot forms a possibly tight curve. The scatter plots for \dslab{} dataset in \figref{scatter_lab} show the difficulty of the task on this dataset --- objective and subjective measures of quality are not well correlated for any of the tested metrics. They correlate even worse on \dsllff{} dataset~\ref{fig:scatter_llff}, which demonstrates that testing on sparse views in current evaluation protocol is insufficient to assess the subjective quality of synthesized videos. The scatter plots, however, form much stronger relations for the \dsfieldwork{} dataset in \figref{scatter_outdoor}. 

\begin{figure*}[t]
\begin{center}
    \includegraphics[width=0.99\textwidth,height=0.5\textwidth]{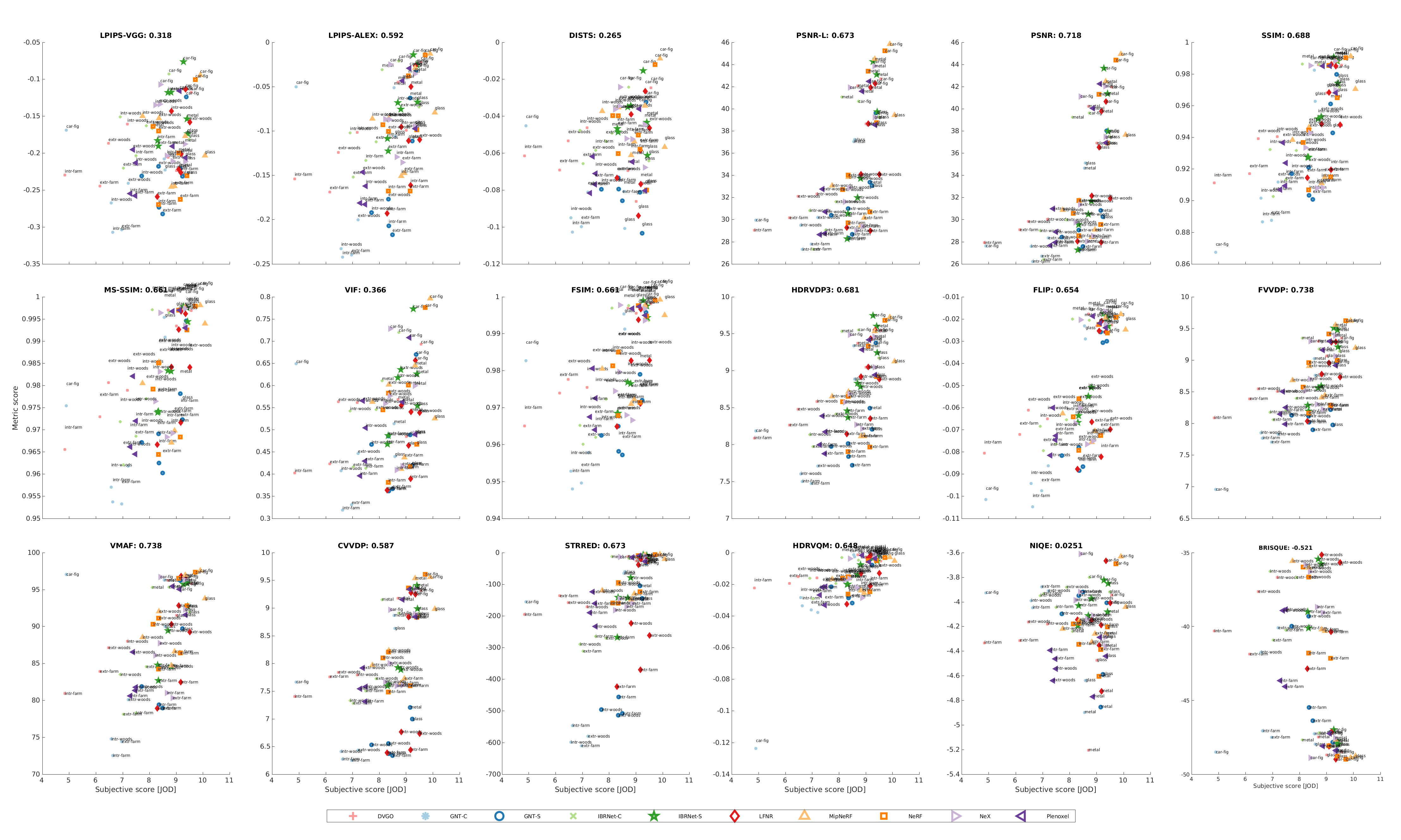}
\end{center}
   \caption{Scatter plots of per-scene metric predictions vs. subjective scores for the \dslab{} dataset. The subjective scores of each scene are shifted such that Reference videos have jod values equal to 10. The numbers above each plot show Spearman correlation. Note that the correlation reported in other plots has been computed on the metric predictions and subjective scores averaged across all scenes. }
\label{fig:scatter_lab}
\end{figure*}

\begin{figure*}[t]
\begin{center}
\includegraphics[width=0.99\textwidth,height=0.55\textwidth]{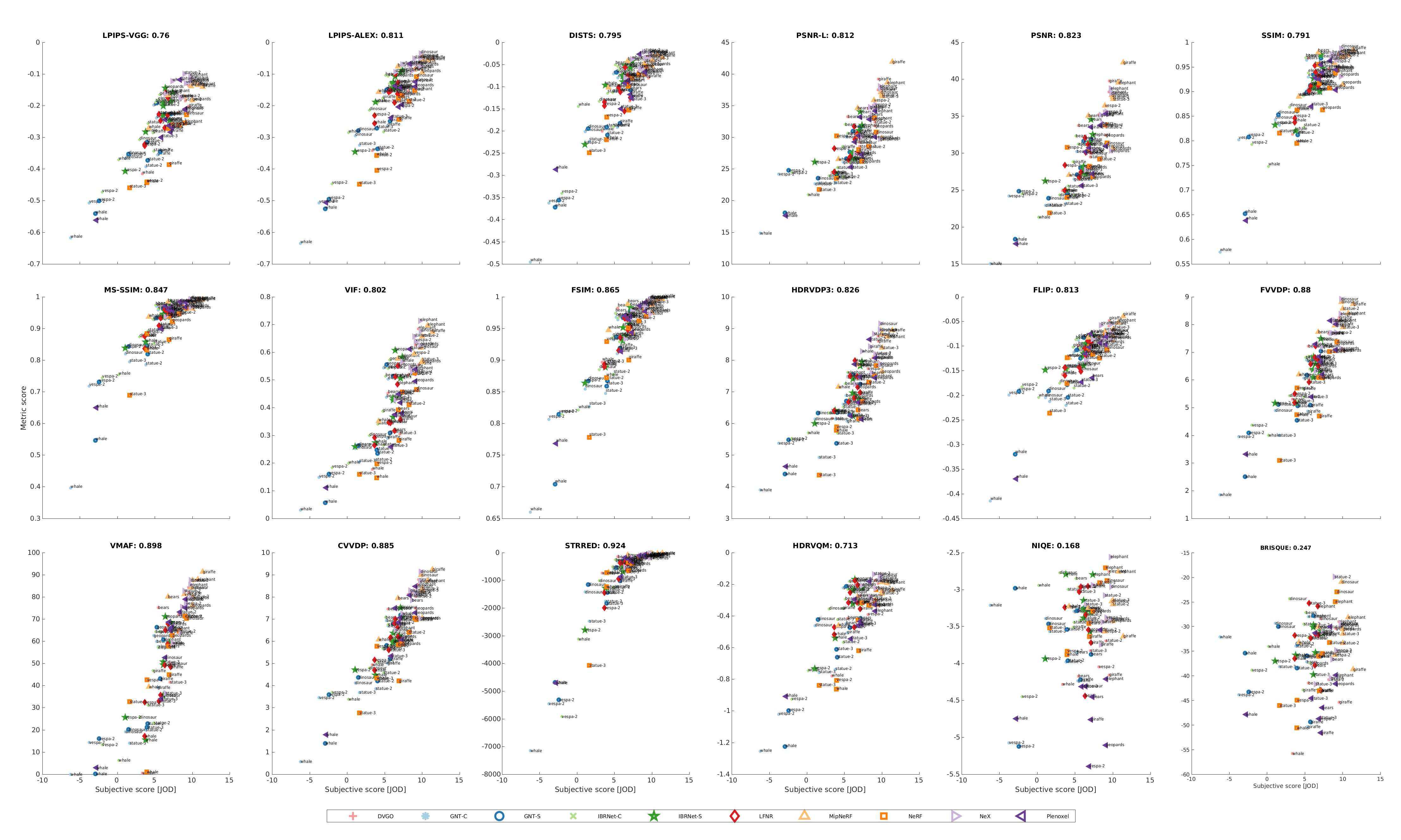}
\end{center}
   \caption{Scatter plots of per-scene metric predictions vs. subjective scores for the \dsfieldwork{} dataset. The notation is the same as in \figref{scatter_lab}.}
\label{fig:scatter_outdoor}
\end{figure*}

\begin{figure*}[t]
\begin{center}
    \includegraphics[width=0.95\textwidth]{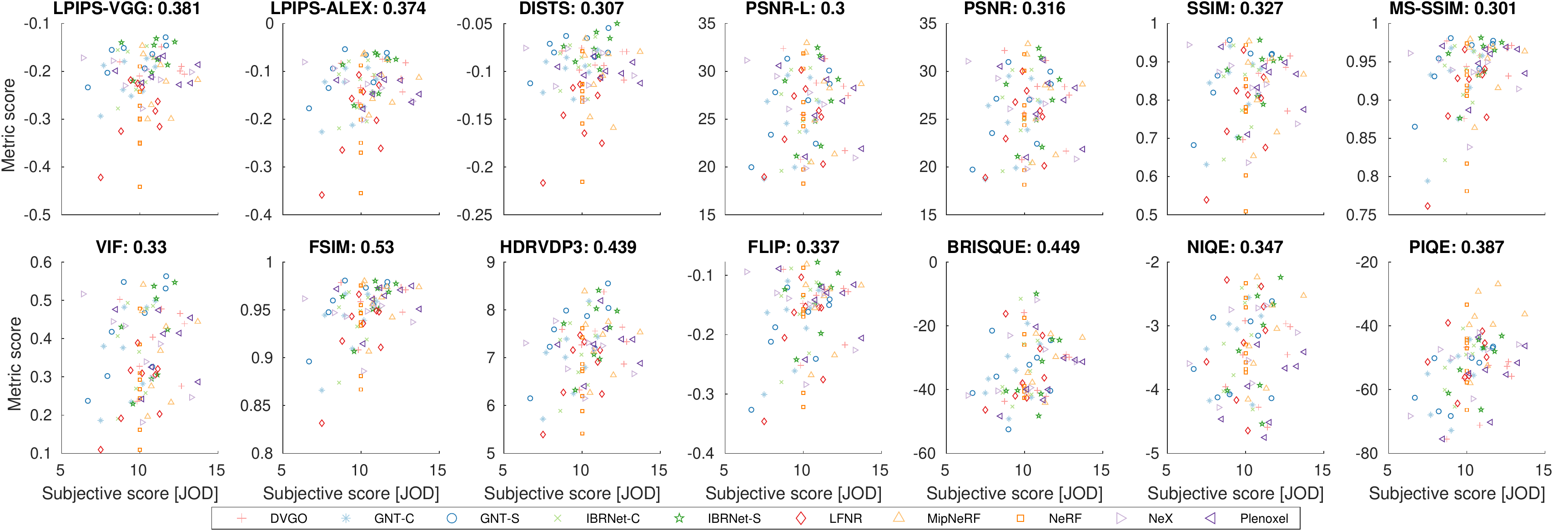}
\end{center}
   \caption{Scatter plots of per-scene metric predictions vs. subjective scores for the \dsllff{} dataset. The subjective scores of each scene are shifted such that NeRF results have jod values equal to 10. The notation is the same as in \figref{scatter_lab}.}
\label{fig:scatter_llff}
\end{figure*}

\begin{figure*}[hb]
\begin{center}
\includegraphics[width=1.0\textwidth]{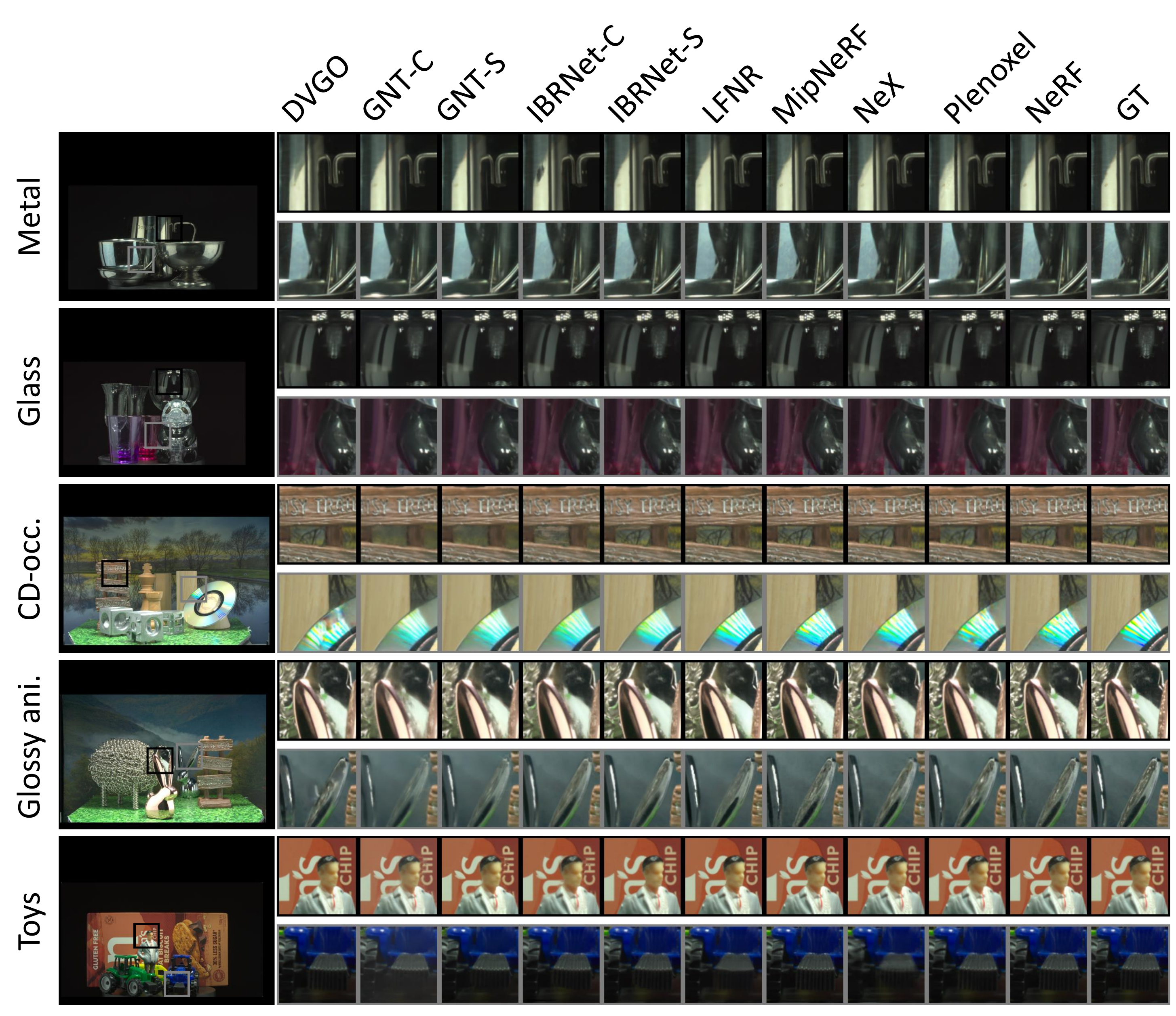}
\end{center}
   \caption{Example inference results for \dslab{} dataset.}
\label{lab_quality}
\end{figure*}

\begin{figure*}[t]
\begin{center}
    \includegraphics[width=\linewidth]{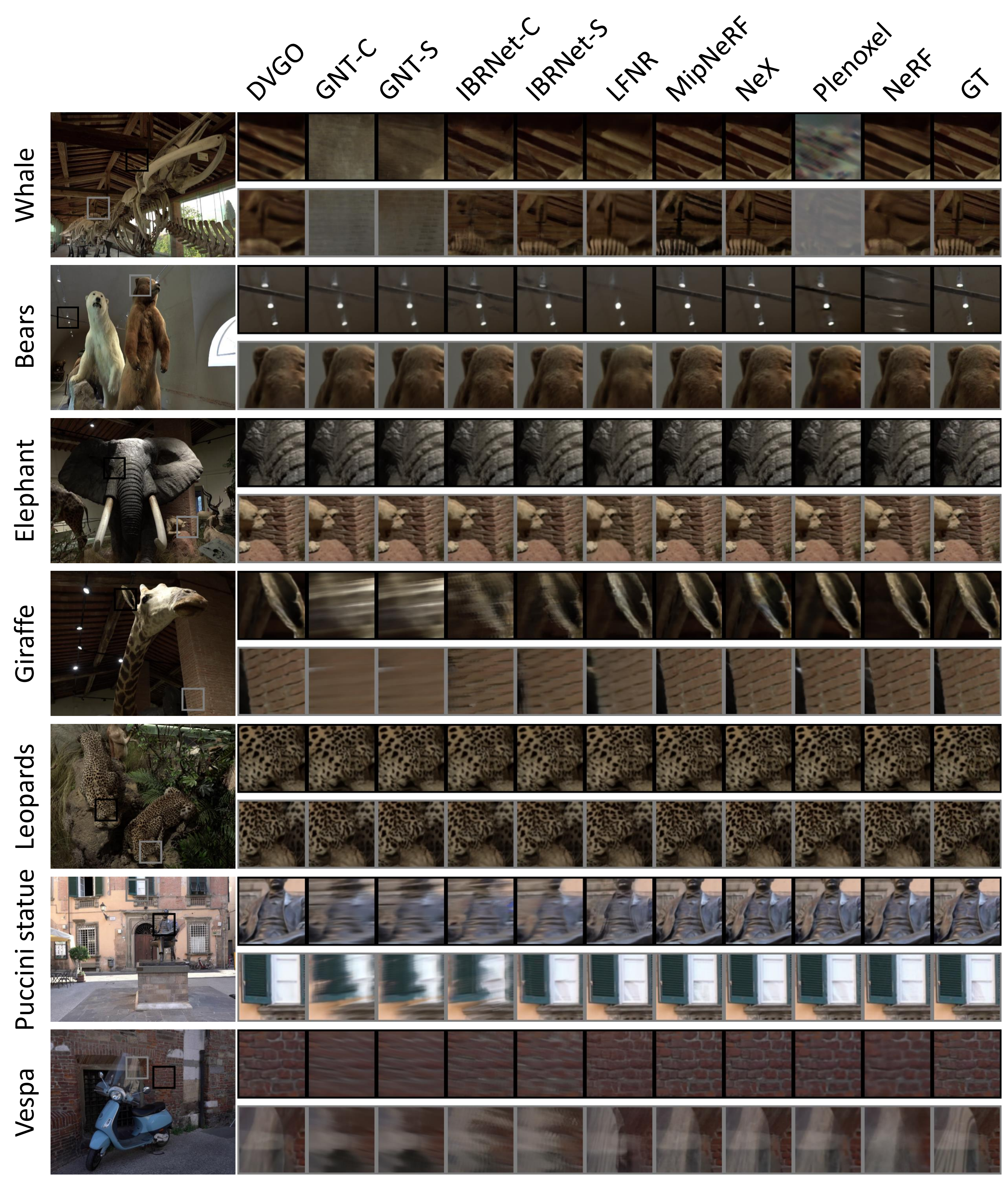}
\end{center}
   \caption{Example inference results for \dsfieldwork{} dataset.}
\label{fieldwork_quality}
\end{figure*}

\section{Training details}
\textbf{DVGO} We follow the training setup as in ~\cite{dirctvoxgo} and set expected numbers of voxels to be $M(c) =
100^3$ and $M(f) = 160^3$ in coarse and fine stages. The points sampling step sizes are set to half
of the voxel sizes, i.e., $\delta(c) = 0.5 \cdot s(c)$ and $\delta(f) = 0.5 \cdot s(f)$.
The shallow MLP layer comprises two hidden layers with
128 channels. The Adam optimizer~\cite{adam} is employed with a batch
size of 8192 rays to optimize the coarse and fine scene representations for 10k and 20k iterations. The base learning
rates are 0.1 for all voxel grids and $10^{-3}$ for the shallow
MLP. The exponential learning rate decay is applied. 

\textbf{NeRF} We follow the pytorch implementation of NeRF~\cite{lin2020nerfpytorch}. We use a batch size of 1024 rays, each sampled at $N_c = 64$
coordinates in the coarse volume and $N_f = 128$ additional coordinates in the
fine volume. We use the Adam optimizer with a base learning rate at
$5 \times 10^{-4}$ and optimize for $2\times10^5$ iterations.

\textbf{GNT} For the cross-scene generalizable GNT model (denoted as GNT-C), we use the pre-trained model released by~\cite{gnt}. For finetuned version of GNT model on each scene (denoted as GNT-S), we finetune the cross-scene model with Adam optimizer with base learning rates for the feature
extraction network and GNT as $10^{-3}$
and $5 \times 10^{-4}$
respectively, which decay exponentially over
training steps. For all our experiments, we train for 50,000 steps with 4096 rays sampled in each
iteration. 

\textbf{IBRNet}
For the cross-scene generaliable IBRNet model (denoted as IBRNet-C), we use the pre-trained model from~\cite{ibrnet}. During fine-tuning stage for IBRNet-S, we optimize both 2D feature extractor and IBRNet itself with Adam optimizer using base learning rates of $5\times 10^{-4}$ and $2\times 10^{-4}$).

\textbf{LFNR}
The architecture of transformer is the same as the ones recently introduced for vision related
tasks \cite{transformer}. Each transformer has 8 blocks and the internal feature size is 256. In each training step, we randomly choose a target image and sample a batch of random
rays from it. The batch size is 128. We train for 250 000 iterations with the Adam optimizer and a linear learning
rate decay schedule with 5000 warm-up steps. 

\textbf{MipNeRF}
We follow the training procedure specified by~\cite{mipnerf}: 1
million iterations of Adam with a batch size of 4096
and a learning rate that is annealed logarithmically from
$5 \cdot 10{-4}$ to $5 \cdot 10^{-6}$.

\textbf{NeX}
As in~\cite{NeX}, we use an
MPI with 192 layers with $ M = 12$  consecutive planes sharing one set of texture coefficients. We sample and render 8,000 pixels in the training view for photometric loss computation. The network is  trained for 4,000 epochs using Adam optimizer with
a learning rate of 0.01 for base color and 0.001 for both
networks and a decay factor of 0.1 every 1,333 epochs.

\textbf{Plenoxel} The implementation of Plenoxel is based on a custom PyTorch CUDA~\cite{cuda} extension library to achieve fast
differentiable volume rendering. We use a batch size of 5000 rays and optimize with RMSProp~\cite{Rmsprop}. For optimization of density, we use the same delayed exponential learning rate schedule as MipNeRF~\cite{mipnerf}, where the exponential is scaled by a learning rate of 30 and decays to
0.05 at step 250000, with an initial delay period of 15000
steps. For SH we adopts a pure exponential decay learning rate
schedule, with an initial learning rate of 0.01 that decays to
$5 \times 10^{-6}$ at step 250000.

\end{document}